\documentclass[10pt,journal,compsoc]{IEEEtran}
\ifCLASSOPTIONcompsoc
  \usepackage[nocompress]{cite}
\else
  \usepackage{cite}
\fi

\newcommand{\etal}{{\emph{et~al.}}~}
\newcommand{\eg}{{\emph{e.g.}}}
\newcommand{\ie}{{\emph{i.e.}}}
\newcommand{\etc}{{\emph{etc}}}

\usepackage{graphicx}
\usepackage{amsmath,amssymb}
\usepackage{bbm}
\usepackage{epsfig}
\usepackage{multirow}
\usepackage{caption}
\usepackage{algorithm}
\usepackage{algpseudocode}
\usepackage{dsfont}
\usepackage{booktabs}
\usepackage{color}
\usepackage{hyperref}

\ifCLASSINFOpdf
\else
\fi
\begin{document}
%
\title{NTU RGB+D 120: A Large-Scale Benchmark \\ for 3D Human Activity Understanding}

\author{Jun~Liu,
        Amir~Shahroudy,
        Mauricio~Perez,
        Gang~Wang, 
        Ling-Yu~Duan, 
        and~Alex~C.~Kot
\IEEEcompsocitemizethanks{\IEEEcompsocthanksitem J. Liu, M. Perez, and A. C. Kot
are with ROSE Lab, School of EEE, Nanyang Technological University, Singapore. \protect\\
E-mail: \{jliu029, mauricio001, eackot\}@ntu.edu.sg.
\IEEEcompsocthanksitem A. Shahroudy is with Department of Electrical Engineering, Chalmers University of Technology, Sweden. \protect\\
E-mail: amirsh@chalmers.se.
\IEEEcompsocthanksitem L.-Y. Duan is with National Engineering Laboratory for Video Technology, Peking University, China,
and also with Peng Cheng Laboratory, China. \protect\\
E-mail: lingyu@pku.edu.cn.
\IEEEcompsocthanksitem G. Wang is with Alibaba Group, China. \protect\\
E-mail: wanggang@ntu.edu.sg.
}
}

\IEEEtitleabstractindextext{%
\begin{abstract}
Research on depth-based human activity analysis achieved outstanding performance and demonstrated the effectiveness of 3D representation for action recognition.
The existing depth-based and RGB+D-based action recognition benchmarks have a number of limitations,
including the lack of large-scale training samples, realistic number of distinct class categories, diversity in camera views, varied environmental conditions, and variety of human subjects.
In this work, we introduce a large-scale dataset for RGB+D human action recognition,
which is collected from 106 distinct subjects and contains more than 114 thousand video samples and 8 million frames.
This dataset contains 120 different action classes including daily, mutual, and health-related activities.
We evaluate the performance of a series of existing 3D activity analysis methods on this dataset,
and show the advantage of applying deep learning methods for 3D-based human action recognition.
Furthermore, we investigate a novel one-shot 3D activity recognition problem on our dataset,
and a simple yet effective Action-Part Semantic Relevance-aware (APSR) framework is proposed for this task,
which yields promising results for recognition of the novel action classes.
We believe the introduction of this large-scale dataset will enable the community to apply, adapt, and develop various data-hungry learning techniques for depth-based and RGB+D-based human activity understanding.
$[$\emph{The dataset is available at: \url{http://rose1.ntu.edu.sg/Datasets/actionRecognition.asp}.}$]$
\end{abstract}

}

\maketitle

\IEEEdisplaynontitleabstractindextext

%
\IEEEpeerreviewmaketitle

\IEEEraisesectionheading{\section{Introduction}\label{sec:intro}}

%
%
%
%

\IEEEPARstart{T}{he} development of depth sensors, \eg, Microsoft Kinect, Intel RealSense, and Asus Xtion,
enables us to obtain effective 3D structure information of the objects and scenes \cite{kinectSurvey2013}.
This empowers the computer vision solutions to move important steps towards 3D vision,
such as 3D object recognition \cite{guo20143d}, 3D scene understanding \cite{gupta2015indoor}, and 3D activity analysis \cite{aggarwal2014survey}, \etc.
Unlike RGB video-based activity analysis \cite{kuehne2011hmdb,soomro2012ucf101,caba2015activitynet,kay2017kinetics},
3D action recognition suffers from the lack of large-scale benchmark datasets.
There are no publicly shared sources like YouTube to supply ``in-the-wild'' 3D video samples of a realistically various set of action classes.
This limits our ability to build large-sized benchmarks to train, evaluate, and compare the strengths of different approaches, especially the recent data-hungry techniques like deep learning methods.
To the best of our knowledge, all the currently available 3D action recognition benchmarks have limitations in various aspects.

First is the small number of subjects and narrow range of performers' ages.
This can significantly limit the intra-class variation of the actions.
The constitution of human activities depend on the gender, age, physical condition, and even cultural aspects of the subjects.
As a result, variation of human subjects is crucial for building a realistic action recognition benchmark.

The second factor is the limited number of action categories.
When only a small set of classes are available, each can be very distinguishable by finding a simple motion pattern or even by the appearance of an interacted object.
However, when the number of classes grows, similar motion patterns and interacted objects will be shared among different classes,
which makes the action recognition much more challenging.

The third limitation is the highly restricted camera views.
In most of the current datasets, the samples are captured from a front view with a fixed camera viewpoint.
In some others, the views are often bounded to fixed front and side views using multiple cameras at the same time.

The fourth factor is the limited variation of the collection environments (\eg, backgrounds) which can also be important to achieve a sensible activity analysis dataset.

Finally and most importantly, the very limited number of video samples hinders the application of the advanced data-driven learning methods to this problem.
Though several attempts have been done \cite{rnnskeleton_cvpr15,cnn_for_depth_action_THMS}, 
they mostly suffer from over-fitting and have to scale down the size of their learning models.
Therefore, they clearly need many more samples to generalize and perform better on the testing videos.

In order to overcome these limitations, a large-scale benchmark dataset,
NTU RGB+D 120 dataset,
is developed for 3D human activity analysis.

\begin{table*}
	\setlength{\tabcolsep}{3pt} 
\caption{Comparison of the proposed NTU RGB+D 120 dataset and some of the other publicly available datasets for 3D action recognition.
		Our dataset provides many more video samples, action classes, human subjects, and camera views in comparison with other available datasets for RGB+D action recognition.}
	\begin{center}
        \footnotesize
		\begin{tabular}{llrrrcllc}
            \toprule
			Datasets			& ~	 			   &~~~~~\#Videos      &\#Classes &\#Subjects &~~~\#Views~~~ &Sensors & Data Modalities		& Year	
            \\ \toprule
			MSR-Action3D		& \cite{msraction3ddataset}	& 567		& 20	& 10		& 1		& N/A	     	& D+3DJoints			& 2010	
            \\ \midrule
			CAD-60				& \cite{cad60}				& 60		& 12	& 4			& -		& Kinect v1 	& RGB+D+3DJoints		& 2011	
            \\ \midrule
			RGBD-HuDaAct		& \cite{rgbdhudaact}		& 1,189		& 13	& 30		& 1		& Kinect v1		& RGB+D					& 2011	
            \\ \midrule
			MSRDailyActivity3D	& \cite{actionletCVPR}		& 320		& 16	& 10		& 1		& Kinect v1		& RGB+D+3DJoints		& 2012	
            \\ \midrule
            UT-Kinect        	& \cite{HOJ3D}	        	& 200		& 10	& 10		& 4		& Kinect v1		& RGB+D+3DJoints		& 2012	
            \\ \midrule
			Act$4^2$			& \cite{Act42}				& 6,844		& 14	& 24		& 4		& Kinect v1		& RGB+D					& 2012	
            \\ \midrule
			CAD-120				& \cite{cad120}				& 120		& 10+10	& 4			& -		& Kinect v1		& RGB+D+3DJoints		& 2013	
            \\ \midrule
			3D Action Pairs		& \cite{HON4D}				& 360		& 12	& 10		& 1		& Kinect v1		& RGB+D+3DJoints		& 2013	
            \\ \midrule
			Multiview 3D Event	& \cite{Multiview3DEvent}	& 3,815		& 8		& 8			& 3		& Kinect v1		& RGB+D+3DJoints		& 2013	
            \\ \midrule
			Northwestern-UCLA	& \cite{NW_UCLA}			& 1,475		& 10	& 10		& 3		& Kinect v1		& RGB+D+3DJoints		& 2014	
            \\ \midrule
			UWA3D Multiview 	& \cite{HOPC}				&$\sim$900	& 30	& 10		& 1		& Kinect v1		& RGB+D+3DJoints		& 2014	
            \\ \midrule
			Office Activity		& \cite{OfficeActivity}		& 1,180		& 20	& 10		& 3		& Kinect v1		& RGB+D					& 2014	
            \\ \midrule
			UTD-MHAD			& \cite{UTD-MHAD}			& 861		& 27	& 8			& 1		& Kinect v1+WIS	& RGB+D+3DJoints+ID		& 2015	
            \\ \midrule
			UWA3D Multiview II	& \cite{HOPC_PAMI}			& 1,075		& 30	& 10		& 5		& Kinect v1		& RGB+D+3DJoints		& 2015	
            \\ \midrule
			M$^2$I        	    & \cite{xu2015multi}		&$\sim$1,800 & 22	& 22		& 2		& Kinect v1		& RGB+D+3DJoints		& 2015	
            \\ \midrule
			SYSU 3DHOI        	& \cite{hu2017jointly}		& 480		& 12	& 40		& 1		& Kinect v1		& RGB+D+3DJoints		& 2017	
            \\ \toprule
            \bf NTU RGB+D 120   & ~                      & \bf 114,480  & \bf 120   & \bf 106   & \bf 155	& \bf Kinect v2	 & \bf RGB+D+3DJoints+IR  & ~
            \\ \bottomrule
		\end{tabular}
	\end{center}
     It is worth mentioning that most of the other datasets were collected on a single or few backgrounds and under fixed illumination condition,
     while there is high variation of environmental conditions in our dataset,
     which uses 96 different backgrounds and contains significant illumination variation.
	\label{tab:datasetcomparison}
\end{table*}

The proposed dataset consists of $114,480$ RGB+D video samples that are captured from 106 distinct human subjects.
We have collected RGB videos, depth sequences, skeleton data (3D locations of 25 major body joints), and infrared frames using Microsoft Kinect v2.
The action samples are captured from 155 different camera viewpoints.
The subjects in this dataset are in a wide range of age distribution (from 10 to 57) 
and from different cultural backgrounds (15 countries),
which brings very realistic variation to the quality of actions.
We also provide the ambiance inconstancy by capturing the dataset under various environmental conditions (96 different backgrounds with illumination variation).

The large amount of variation in subjects, views, and backgrounds makes it possible to have more sensible cross-subject and cross-setup evaluations for various 3D-based action analysis methods.
The proposed dataset
will help the community to move steps forward in 3D human activity analysis,
and make it possible to apply data-hungry methods, such as deep learning techniques, for this task.

The performance of state-of-the-art 3D action recognition approaches is evaluated on our dataset,
which shows the capability of applying deep models for activity analysis with the suggested cross-subject and cross-setup evaluation criteria.
We also evaluate the performance of fusion across different data modalities, \eg, RGB, depth, and skeleton data, for activity analysis,
since they provide appearance and geometrical information respectively, and are complementary for more accurate action recognition.

In this paper, we also investigate a novel one-shot 3D action recognition problem based on the proposed dataset.
An Action-Part Semantic Relevance-aware (APSR) framework is proposed to handle this task
by utilizing the semantic relevance between each body part and each action class at the distributed word embedding level \cite{zamani2017relevance,mikolov2013distributed}.
Human actions can be represented by a combination of the movements of different body parts \cite{ye2013survey,liu2017global_CVPR},
and the importance degrees of body parts' motion are not equal for recognizing different action categories.
As a result, we need to put more emphasis on the more relevant body parts when recognizing an action performed by a human.
In this paper, we show that the name (text description) of the novel action class can assist in the identification of the relevant body parts,
and by exploiting the semantic relevance between the action's and body part's descriptions as a guidance,
the relevant body parts of the novel action categories can be emphasized,
and thus the one-shot recognition performance is improved.

The rest of this paper is organized as follows.
Section \ref{sec:relatedwork} reviews the current 3D-based human action recognition approaches and benchmarks.
Section \ref{sec:dataset} introduces the proposed dataset, its structure, and the defined evaluation criteria.
Section \ref{sec:approach} presents the proposed APSR framework for one-shot 3D human action recognition.
Section \ref{sec:exp} shows the experimental evaluations with our benchmark.
Finally, section \ref{sec:conclusion} concludes the paper.


\section{Related work}
\label{sec:relatedwork}

We briefly review publicly available benchmark datasets and recent methods for 3D human activity analysis in this section.
Here we introduce a limited number of the most famous ones.
Readers are referred to these survey papers
\cite{RGB-D_Survey_40_Pichao,wang2018rgb,presti20163d,Chen20131995,han2017space,IJPRAI_Survey_Kinect}
for a more extensive list of the current 3D activity analysis datasets and methods.

\subsection{3D Activity Analysis Datasets}
After the release of the Microsoft Kinect \cite{kinectSurvey2012},
several datasets have been collected to conduct research on 3D human action recognition and to evaluate different methods in this field.

The MSR-Action3D dataset \cite{msraction3ddataset} was the earliest which opened up the research in depth-based action analysis.
The samples of this dataset were limited to depth sequences of gaming actions,
\eg, \emph{forward punching, side boxing, forward kicking, side kicking, tennis swinging, tennis serving, golf swinging,} \etc.
Later, the skeleton data was added to this dataset.
The skeletal information includes the 3D locations of 20 different joints at each frame.
A decent number of methods have been evaluated on this benchmark,
and the recent ones reported close to saturation accuracies \cite{MMMP_PAMI,tanfous2018coding}.

The MSR-DailyActivity dataset \cite{actionletCVPR} was among the most challenging benchmarks in this field.
It contains 320 samples of 16 daily activities with higher intra-class variation.
The small number of samples and fixed camera viewpoints are the limitations of this dataset.
Some reported results on this dataset also achieved very high accuracies \cite{RangeSample, jianfang_CVPR15, Luo_2013_ICCV,DSSCA-PAMI}.

The RGBD-HuDaAct dataset \cite{rgbdhudaact} was one of the largest datasets.
It contains RGB and depth sequences of 1189 videos of 12 human daily actions (plus one background class), with high variation in time lengths.
The special characteristic of this dataset is the synced and aligned RGB and depth channels,
which enables local multi-modal analysis of RBGD
\footnote{
We emphasize the difference between RGBD and RGB+D terms.
We suggest using RGBD when the two modalities are aligned pixel-wise, and RGB+D when the resolutions of the two are different and frames are not aligned.}
signals.

The CAD-60 \cite{cad60} and CAD-120 \cite{cad120} datasets contain RGB, depth, and skeleton data of human actions.
The special characteristic of these datasets is the variety of camera views.
Unlike most of the other datasets, the cameras in these two datasets were not bound to front-view or side-views.
However, the limited number of video samples (60 and 120) is the downside of them.

The 3D Action Pairs dataset \cite{HON4D} was proposed to provide multiple pairs of action classes.
Each pair contains very closely related actions with differences along temporal axis,
\eg, \emph{pick up/put down a box, push/pull a chair, put on/take off a hat, etc}.
State-of-the-art methods \cite{Kong_2015_CVPR,MMMP_PAMI,DSSCA-PAMI} achieved perfect accuracy on this benchmark.

The Northwestern-UCLA \cite{NW_UCLA} and the Multiview 3D Event \cite{Multiview3DEvent} datasets used more than one depth sensors at the same time
to collect multi-view representations of the same action, and scale up the number of samples.

The G3D \cite{bloom2012g3d} and PKUMMD \cite{liu2017pku} datasets were introduced for activity analysis in continuous sequences,
which respectively contain 210 and 1076 videos.
In each dataset, the videos were collected in the same environment.

It is worth mentioning that there are more than 40 datasets for 3D human action recognition \cite{RGB-D_Survey_40_Pichao}.
Though each of them provided important challenges of human activity analysis, they have limitations in some aspects.
\tablename{~\ref{tab:datasetcomparison}} shows the comparison between some of the current datasets and our large-scale RGB+D action recognition dataset.

By summarizing the contributions of our dataset over the existing ones, NTU RGB+D 120 dataset has:
(1) many more action classes;
(2) many more video samples for each action class;
(3) much more intra-class variation, \eg, poses, interacted objects, ages and cultural backgrounds of the actors;
(4) many more collection environments, \eg, different backgrounds and illumination conditions;
(5) more camera views;
(6) more camera-to-subject distance variation;
(7) used Kinect v2 which provides more accurate depth-maps and 3D joints, especially in a multi-camera setup, compared to the previous version of Kinect.

This work is an extension of our previous conference paper \cite{Amir-Dataset-CVPR}.
In \cite{Amir-Dataset-CVPR}, we introduced the preliminary version of our dataset that contains 60 action classes.
In this paper, we significantly extend it and build the NTU RGB+D 120 dataset
that is much larger and provides much more variation of environmental conditions, subjects, and camera views, \etc.
It also provides more challenges of 3D human activity analysis.
A brief comparison between these two versions is shown in \tablename{~\ref{tab:twoVersionComparion}}.
Besides, in this paper, we propose a new framework for one-shot 3D action recognition based on the proposed NTU RGB+D 120 dataset.

\begin{table}
    	\caption{Comparison of the dataset version introduced in this paper and the one released in our preliminary conference paper \cite{Amir-Dataset-CVPR}.
                The top three rows show a comparison of the sizes of these two versions.
                The bottom two rows show a comparison of the recognition accuracies when evaluating several methods on it.
                The methods \cite{Liu_2016_ECCV,ke2017new_CVPR} are evaluated by using the suggested two evaluation criteria.}
	\begin{center}
        \small
		\begin{tabular}{lcc}
			\toprule
            Dataset                             & Preliminary \cite{Amir-Dataset-CVPR} &  Extended  \\
            Version                             &   (NTU RGB+D 60)                         & (NTU RGB+D 120)  \\
            \toprule
			\#Videos                            &   56,880            & 114,480 \\ \midrule
            \#Classes                           &   60                & 120     \\ \midrule
            \#Subjects                          &   40                & 106     \\ \midrule \midrule
            ST-LSTM \cite{Liu_2016_ECCV}        &  69.2\%~~77.7\%     &  55.7\%~~57.9\%  \\ \midrule
            MTLN \cite{ke2017new_CVPR}          &  79.6\%~~84.8\%     &  58.4\%~~57.9\%  \\
            \bottomrule
		\end{tabular}
	\end{center}
	\label{tab:twoVersionComparion}
\end{table}

\subsection{3D Action Recognition Methods}

{\bf 3D action recognition by hand-crafted features. }
Since the introduction of the first few benchmarks,
such as the MSR-Action3D \cite{msraction3ddataset} and MSR-DailyActivity \cite{actionletCVPR} datasets,
a decent number of feature extraction and classifier learning methods have been proposed and evaluated based on them.

Oreifej and Liu \cite{HON4D} proposed to calculate the four-dimensional normals (X-Y-depth-time) from depth sequences,
and accumulate them on spatio-temporal cubes as quantized histograms over 120 vertices of a regular polychoron.
Yang and Tian \cite{Yang_2014_CVPR} proposed supernormal vectors as aggregated dictionary-based codewords of four-dimensional normals over space-time grids.
The work of \cite{HOPC_PAMI} introduced histograms of oriented principle components of depth cloud points to extract robust features against viewpoint variations.
Lu \etal \cite{RangeSample} applied $\tau$ test-based binary range-sample features on depth maps and achieved robust representation against noise, scaling, camera views, and background clutter.

To have view-invariant representations of the actions, features can be extracted from the 3D body joint positions \cite{shotton2011real,liu2019feature} which are available for each frame.
Evangelidis \etal \cite{skeletalQuads} divided the body into part-based joint quadruples, and encoded the configuration of each part with a succinct 6D feature vector, so called skeletal quads.
To aggregate the skeletal quads, they applied Fisher vectors, and classified the samples by a linear SVM.
In \cite{VemulapalliCVPR14}, different skeleton configurations were represented as points on a Lie group.
Actions as time-series of skeletal configurations were encoded as curves on this manifold.
The work of \cite{Luo_2013_ICCV} utilized group sparsity-based class-specific dictionary coding with geometric constraints to extract skeleton features.

In most of the 3D action recognition scenarios, there are more than one modality of information,
and combining different data modalities can help to improve the classification accuracy.
Ohn-Bar and Trivedi \cite{hog2-ohnbar} combined second order joint-angle similarity representations of skeletons with a modified two step HOG feature on spatio-temporal depth maps to build global representation of each video sample, and utilized a linear SVM to classify the actions.
Wang \etal \cite{actionletPAMI} combined Fourier temporal pyramids of skeletal information with local occupancy pattern features extracted from depth maps,
and applied a data mining framework to discover the most discriminative combinations of body joints.
A structured sparsity-based multi-modal feature fusion technique was introduced by \cite{AmirAthens} for action recognition in RGB+D domain.
In \cite{6836044}, random decision forests were utilized for learning and feature pruning over a combination of depth and skeleton-based features.
Hu \etal \cite{jianfang_CVPR15} introduced a joint heterogeneous feature learning framework by combing RGB, depth, and skeleton data for human activity recognition.
The work of \cite{MMMP_PAMI} proposed hierarchical mixed norms to fuse different features and select most informative body parts in a joint learning framework.

{\bf 3D action recognition with deep networks.}
Recently, deep learning based-approaches have been proposed for 3D human activity analysis \cite{diffRNN, rnnskeleton_cvpr15, cooccurrance}.
Specifically, many of them have been evaluated based on the preliminary version \cite{Amir-Dataset-CVPR} of our dataset,
or pre-trained on it for transfer learning for other tasks
\cite{luo2017unsupervised,huang2017deep,zolfaghari2017chained,DSSCA-PAMI,keglobal,
kim2017interpretable,rahmani2017learning,luvizon20182d,cavazza2017kernel,baradel2017pose,
baradel2018glimpse,wang2018video,hu2018deep,zhang2018fusing,zhang2019real,ke2018computer,liu20173d,
tang2018deep,wang2017modeling,shi2017learning}.

%


Some approaches used recurrent neural networks (RNNs) to model the motion dynamics and context dependencies for 3D human action recognition.
Du \etal \cite{rnnskeleton_cvpr15} proposed a multi-layer RNN framework for 3D action recognition based on a hierarchy of skeleton-based inputs.
Liu \etal \cite{Liu_2016_ECCV} introduced a Spatio-Temporal LSTM network by modeling the context information in both temporal and spatial dimensions.
Zhang \etal \cite{zhang2017view} added a view-adaptation scheme to the LSTM network to regulate the observation viewpoints.
Luo \etal \cite{luo2017unsupervised} proposed an unsupervised learning method by using a LSTM encoder-decoder framework for action recognition in RGB+D videos.

Convolutional neural networks (CNNs) have also been applied to 3D human action recognition.
Wang \etal \cite{wang2017scene} proposed a ``scene flow to action map'' representation for RGB+D based action recognition with CNNs.
Ke \etal \cite{ke2017skeletonnet} transformed the 3D skeleton data to ten feature arrays and input them to CNNs for action recognition.
Rahmani \etal \cite{rahmani2017learning} designed a deep CNN model
to transfer the visual appearance of human body-parts acquired from different views to a view-invariant space for depth-based activity analysis.

Beside RNNs and CNNs, some other deep models have also been introduced for 3D human action recognition.
Huang \etal \cite{huang2017deep} incorporated Lie group structure into a deep architecture for skeleton-based action recognition.
Tang \etal \cite{tang2018deep} applied deep progressive reinforcement learning to distil the informative frames in the video sequences.
Rahmani and Mian \cite{Rahmani_2015_CVPR} introduced a nonlinear knowledge transfer model to transform different views of the human actions to a canonical view for action classification.

{\bf One-shot 3D Action Recognition. }
Plenty of advanced techniques, such as metric learning \cite{koch2015siamese} and meta learning \cite{ravi2017optimization,yang2018one},
have been introduced for one-shot object recognition and image classification \cite{fei2006one,wang2017multi,vinyals2016matching}.
In the context of 3D activity analysis, there are also a few attempts on one-shot-based learning.

Fanello \etal \cite{fanello2013one} used 3D-HOF and Global-HOG features for one-shot action recognition in RGB+D videos.
Wan \etal \cite{wan2016explore} extracted mixed features around the sparse keypoints that are robust to scale, rotation and partial occlusions.
Kone{\v{c}}n{\`y} \etal \cite{konevcny2014one} combined HOG and HOF descriptors together with the dynamic time warping technique for one-shot RGB+D-based action recognition.

Different from these works,
a simple yet effective APSR framework is introduced in this paper,
which emphasizes the features of the relevant body parts by considering the semantic relevance of the action class and each body part,
for one-shot 3D action recognition.


\begin{figure}
	\centering
	\includegraphics[width=0.82\linewidth]{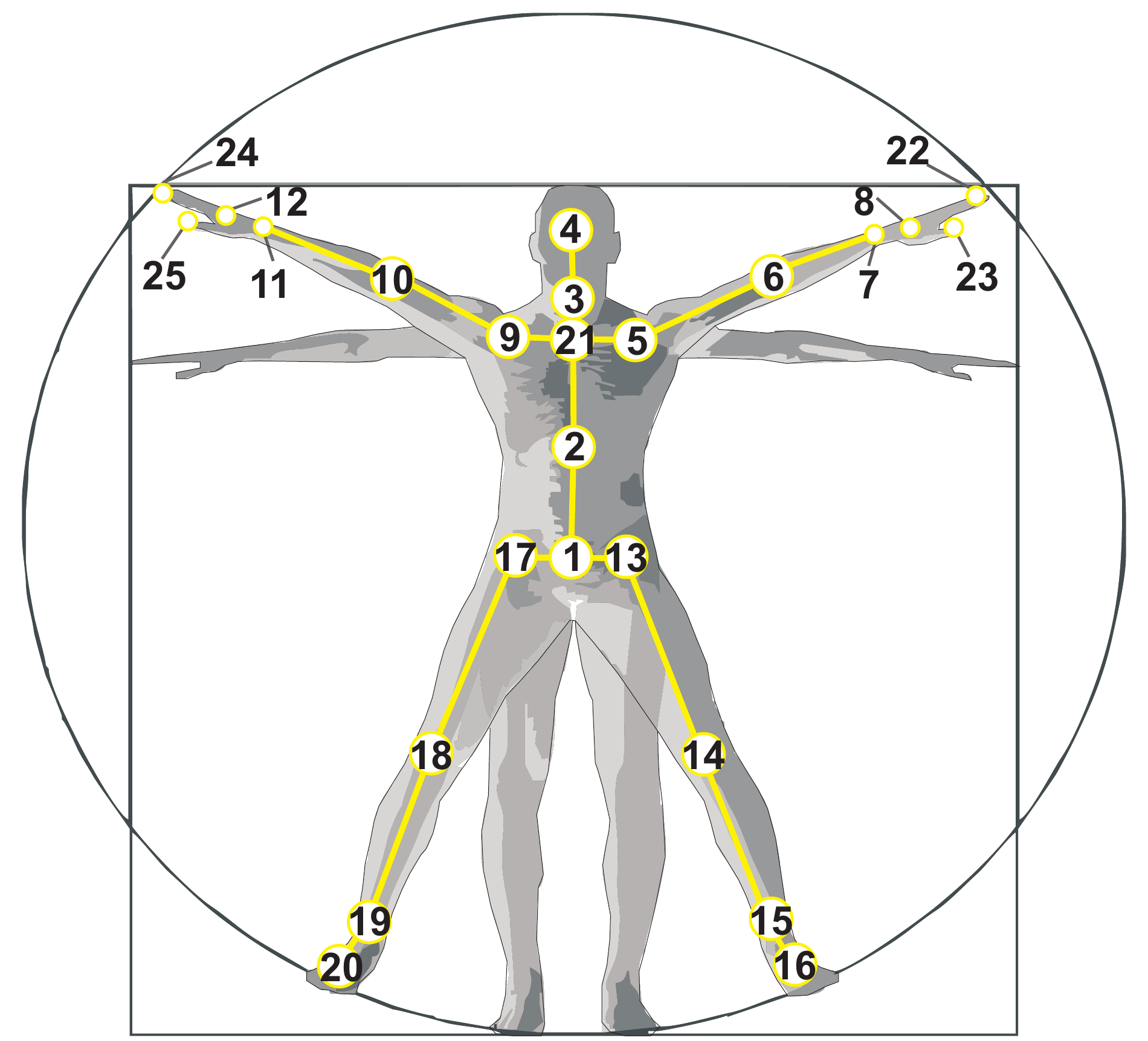}
	\caption{Illustration of the configuration of 25 body joints in our dataset.
		The labels of these joints are:
		(1) base~of~spine,
		(2) middle~of~spine,
		(3) neck,
		(4) head,
		(5) left~shoulder,
		(6) left~elbow,
		(7) left~wrist,
		(8) left~hand,
		(9) right~shoulder,
		(10) right~elbow,
		(11) right~wrist,
		(12) right~hand,
		(13) left~hip,
		(14) left~knee,
		(15) left~ankle,
		(16) left~foot,
		(17) right~hip,
		(18) right~knee,
		(19) right~ankle,
		(20) right~foot,
		(21) spine,
		(22) tip~of~left~hand,
		(23) left~thumb,
		(24) tip~of~right~hand,
		(25) right~thumb.
	}
	\label{fig:skeleton}
\end{figure}

\section{The NTU RGB+D 120 Dataset}
\label{sec:dataset}

In this section, we introduce the details of the proposed NTU RGB+D 120 action recognition dataset and the defined evaluation criteria.

\subsection{Dataset Structure}
\label{sec:datasetstructure}

\subsubsection{Data Modalities}
We use Microsoft Kinect sensors to collect our dataset.
We collect four major data modalities acquired by this sensor,
namely, the depth maps, the 3D joint information, the RGB frames, and the infrared (IR) sequences.

The depth maps are sequences of two dimensional depth values in millimeters.
To maintain all the information, we apply lossless compression for each individual frame.
The resolution of each depth frame is $512\times424$.

The joint information consists of 3-dimensional locations of 25 major body joints for each detected and tracked human body in the scene.
The corresponding pixels on RGB frames and depth maps are also provided for each body joint.
The configuration of these joints is illustrated in \figurename{~\ref{fig:skeleton}}.

The RGB videos are recorded in the provided resolution of $1920\times1080$.

The infrared sequences are also collected and stored frame by frame at the resolution of $512\times424$.

\subsubsection{Action Classes}
We have 120 action categories in total, which are divided into three major groups, including
82 daily actions (eating, writing, sitting down, moving objects, \etc),
12 health-related actions (blowing nose, vomiting, staggering, falling down, \etc),
and 26 mutual actions (handshaking, pushing, hitting, hugging, \etc).

Compared to the preliminary version \cite{Amir-Dataset-CVPR} of our dataset,
the proposed NTU RGB+D 120 dataset contains many more action classes.
Here we summarise the characteristics of the newly added actions compared to the actions in the preliminary version:
\emph{(1) Fine-grained hand/finger motions.}
Most of the actions in the preliminary version of our dataset have significant body and hand motions,
while the newly added classes in this extended version contain some actions that have fine-grained hand and finger motions,
such as ``make ok sign'' and ``snapping fingers''.
\emph{(2) Fine-grained object-related individual actions.}
The newly added actions include some fine-grained object-involved single-person actions,
in which the body movements are not significant and the sizes of the involved objects are relatively small,
such as ``counting money'' and ``play magic cube''.
\emph{(3) Object-related mutual actions.}
Most of the two-person mutual actions in the preliminary version do not involve objects.
In this extended version, some of the newly added mutual actions involve the interactions with objects,
such as ``wield knife towards other person'' and ``hit other person with object''.
\emph{(4) Different actions with similar posture patters but with different motion speeds.}
In this extended version, there are some different actions that have similar posture patterns but have different motion speeds.
For example, ``grab other person's stuff'' is a newly added action,
and its main difference compared to ``touch other person's pocket (steal)'' is the motion speed.
\emph{(5) Different actions with similar body motions but with different objects involved.}
There are some different actions that have very similar body motions but involve different objects.
For example, the motions in the newly added action ``put on bag/backpack'' are similar to those in the existing action ``put on jacket''.
\emph{(6) Different actions with similar objects involved but with different body motions.}
Among the newly added actions, there are also some different classes that share the same interacted objects, such as ``put on bag/backpack'' and ``take something out of a bag/backpack''.

\subsubsection{Subjects}
We invited 106 distinct subjects to our dataset collection sessions.
These subjects are from 15 different countries.
Their ages are between $10$ and $57$, and heights are between $1.3m$ and $1.9m$.
\figurename{~\ref{fig:sampleframes}} illustrates the variety of the subjects in age, gender, and height.
Each subject is assigned a consistent ID number over the entire dataset.

%

\subsubsection{Collection Setups}
We use 32 collection setups to build our dataset,
and over different setups, we change the location and background.
%
Specifically, in each setup,
we use three cameras at the same time to capture three different horizontal views for the same action sample.
The three cameras are set up at the same height yet from three different horizontal angles: $-45^\circ, 0^\circ, +45^\circ$.
Each subject is asked to perform each action twice, once towards the left camera, and once towards the right camera.
In this way, in each collection setup, we capture two front views, one left side view,
one right side view, one left side $45$ degrees view, and one right side $45$ degrees view. 
The three cameras are assigned consistent camera numbers in our dataset, where
camera $1$ always observes the $45$ degrees views, while cameras $2$ and $3$ observe the front and side views.

To further increase the number of camera views,
over different collection setups, we also change the vertical heights of the cameras and their distances to the subjects,
as reported in \tablename{~\ref{tab:camerasetups}}.
All the camera and setup numbers are provided for each video sample in our dataset.

\subsection{Benchmark Evaluations}
\label{sec:evaluations}

To have standard evaluations for the methods to be tested on this benchmark,
we define precise criteria for two types of action classification evaluation.
For each criterion, we report the classification accuracy in percentage.

\subsubsection{Cross-Subject Evaluation}
\label{sec:crosssubjevaluation}
For cross-subject evaluation, the 106 subjects are split into training and testing groups.
Each group consists of 53 subjects.
The IDs of the training subjects in this evaluation are:
1, 2, 4, 5, 8, 9, 13, 14, 15, 16, 17, 18, 19, 25, 27, 28, 31, 34, 35, 38, 45, 46, 47,
49, 50, 52, 53, 54, 55, 56, 57, 58, 59, 70, 74, 78, 80, 81, 82, 83, 84, 85, 86, 89, 91, 92, 93, 94, 95, 97, 98, 100, 103.
The remaining subjects are reserved for testing.

\subsubsection{Cross-Setup Evaluation}
\label{sec:crosssubjevaluation}
For cross-setup evaluation, we pick all the samples with even collection setup IDs for training, and those with odd setup IDs for testing,
\ie, 16 setups are used for training, and the other 16 setups are reserved for testing.

\begin{table}
	\setlength{\tabcolsep}{4pt} 
	\caption{The cameras' height and distance to the subjects in each collection setup.
        }
	\begin{center}
      \footnotesize
		\begin{tabular}{cccccc}
			\toprule
			Setup  & Height     & Distance	 & Setup  & Height     & Distance   \\
			No.    & (\emph{m}) & (\emph{m}) & No.    & (\emph{m}) & (\emph{m}) \\
            \midrule  
			(01) & 1.7 & 3.5 & (02) & 1.7 & 2.5 \\ 
			(03) & 1.4 & 2.5 & (04) & 1.2 & 3.0 \\
			(05) & 1.2 & 3.0 & (06) & 0.8 & 3.5 \\
			(07) & 0.5 & 4.5 & (08) & 1.4 & 3.5 \\
			(09) & 0.8 & 2.0 & (10) & 1.8 & 3.0 \\
			(11) & 1.9 & 3.0 & (12) & 2.0 & 3.0 \\
			(13) & 2.1 & 3.0 & (14) & 2.2 & 3.0 \\
			(15) & 2.3 & 3.5 & (16) & 2.7 & 3.5 \\
			(17) & 2.5 & 3.0 & (18) & 1.8 & 3.3 \\
            (19) & 1.6 & 3.5 & (20) & 1.4 & 4.0 \\
            (21) & 1.7 & 3.2 & (22) & 1.9 & 3.4 \\
            (23) & 2.0 & 3.2 & (24) & 2.4 & 3.3 \\
            (25) & 2.5 & 3.3 & (26) & 1.5 & 2.7 \\
            (27) & 1.3 & 3.5 & (28) & 1.1 & 2.9 \\
            (29) & 2.5 & 2.8 & (30) & 2.4 & 2.7 \\
            (31) & 1.6 & 3.0 & (32) & 2.3 & 3.0 \\
            \bottomrule
		\end{tabular}
	\end{center}
	\label{tab:camerasetups}
\end{table}

\section{APSR Framework for One-Shot 3D Action Recognition}
\label{sec:approach}
Existing works \cite{fei2006one,wang2017multi} show that once some categories have been learned,
the knowledge gained in this process can be abstracted and used to learn novel classes efficiently,
even if only one learning example per new class is given (\ie, via one-shot learning).
Since the samples of certain categories may be difficult to collect \cite{wang2018rgb},
one-shot visual recognition becomes an important research branch in computer vision.

In this section, we introduce the one-shot 3D action recognition scenario based on our proposed dataset,
and show how a large auxiliary set could be used to assist the one-shot recognition for the novel classes.
Specifically, an Action-Part Semantic Relevance-aware (APSR) framework is proposed for more reliable one-shot 3D action recognition.

\subsection{One-Shot Recognition on NTU RGB+D 120}
\label{sec:approach:oneshot}

We define the one-shot 3D action recognition scenario on the proposed NTU RGB+D 120 dataset as follows.

We split the full dataset into two parts: the auxiliary set and the evaluation set.
There are no overlaps of classes between these two sets.
The auxiliary set contains multiple action classes and samples,
and these samples can be used for learning (\eg, learning a feature generation network).
The evaluation set consists of the novel action classes for one-shot recognition evaluation,
and one sample from each novel class is picked as the exemplar, while the remaining samples are used to test the recognition performance.

\subsection{APSR Framework}
\label{sec:approach:semanticawarenet}

Previous works \cite{chen2016novel,liu2017global_CVPR} have shown that
the importance degrees of the features from different body parts are not the same in analyzing different actions.
For example, the features extracted from the leg joints are more relevant in recognizing the action ``kicking'', compared to those from other body parts.
Therefore, it is intuitive to identify the body parts that are more relevant to the action performed in a video sequence,
and correspondingly emphasize their features to achieve reliable recognition performance.
%

However, in our one-shot recognition scenario, identifying the relevant body parts of the novel actions is difficult,
as learning to generalize beyond the single specific exemplar is often very challenging \cite{wang2017multi}.

In this paper, a simple yet effective APSR framework,
which can be used to generalize to the novel action categories, is introduced for one-shot recognition.
The APSR framework emphasizes the relevant body parts for each new class of actions by considering the semantic relevance between the action and each body part based on their descriptions.
Specifically, we design a network to generate the features of each body part,
and then perform weighted pooling over these features with the guidance of relevance scores in the word embedding space.
Finally, the pooling result is used to perform one-shot recognition.

\textbf{Feature Generation Network.}
The feature generation network is adopted from the 2D Spatio-Temporal LSTM (ST-LSTM) \cite{liu2017skeleton_PAMI} designed for 3D activity analysis,
which models the context and dependency information in both temporal dimension (over different frames) and spatial dimension (over different body parts).
Readers are referred to \cite{liu2017skeleton_PAMI} for more details about the mechanism of ST-LSTM.

\begin{figure}
	\centering
	\includegraphics[width=0.6\linewidth]{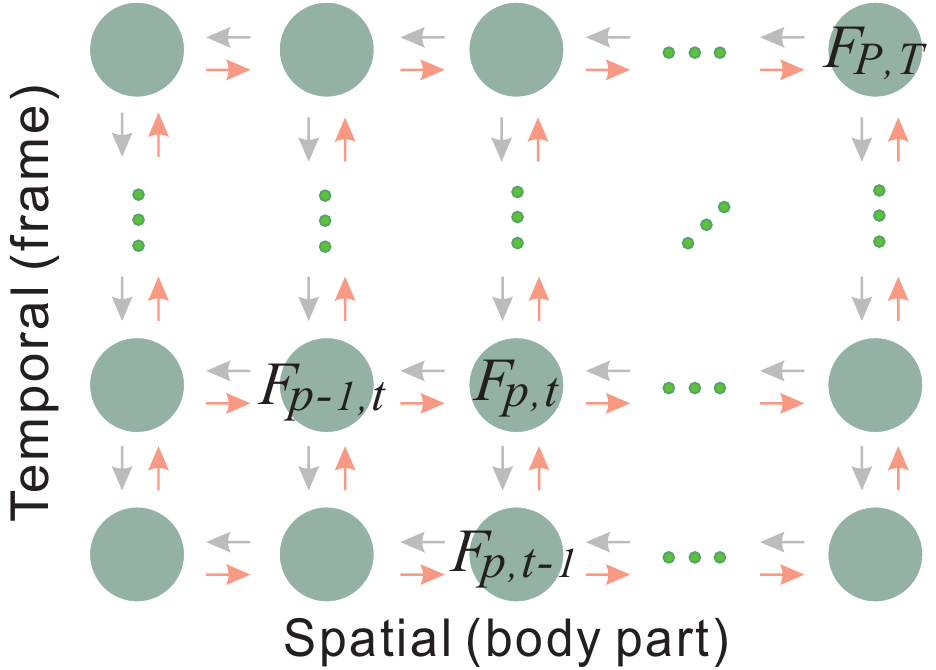}
	\caption{Illustration of the body part feature generation network.
	}
	\label{fig:stlstm}
\end{figure}

The original ST-LSTM \cite{liu2017skeleton_PAMI} models the context information via a single pass.
To produce a more discriminative set of features for each part,
we introduce bidirectional context passing (similar to bidirectional LSTM \cite{graves2013hybrid})
for our feature generation network, as illustrated in \figurename{~\ref{fig:stlstm}}.

The input to our feature generation network is an action sample (here we input its skeleton data for efficiency),
and the outputs are the features of all body parts at each frame.
Concretely,
at the unit $(p,t)$ of this network, the input is the 3D coordinates of the skeletal joint of the body part ($p$) in the frame ($t$),
and the output is the feature ($F_{p,t}$) representing this body part at this frame.
By incorporating the spatio-temporal context information into each part,
the obtained feature set $\mathbf{F} = \{F_{p,t} ~ | ~ p\in\{1,...,P\}, t\in\{1,...,T\} \}$
is powerful for representing each body part ($p$) at each frame ($t$) in the performing of the action instance,
where $P$ is the number of body parts, 
and $T$ is the number of frames used for each video sample.

\textbf{Semantic Relevance.}
The proposed method is inspired by the recent works on word embedding in natural language processing \cite{mikolov2013distributed,zamani2017relevance,pennington2014glove}.
In these works, each word is mapped into a continuous embedding space,
and two semantically relevant words will have large cosine similarity in this space \cite{pennington2014glove,de2016representation,li2017leveraging}.
By pre-training on a massive natural language corpus,
these models demonstrate their superior ability in preserving the semantic relations among different words,
and thus have been successfully transferred to different tasks,
such as document classification \cite{yang2016hierarchical}, image classification\cite{wang2017multi}, and image/video caption generation \cite{karpathy2015deep}.

This motivates us to utilize the prior knowledge about relevant body parts for recognizing new action classes,
and the semantic relevance (cosine similarity) between the novel action's name and each body part's name in the embedding space can be used as prior information.

\begin{figure}[tb]
    \begin{minipage}[h]{1.0\linewidth}
		\centering
		\centerline{\includegraphics[scale=0.399,trim={1.0cm 6.6cm 1.8cm 4.65cm},clip]{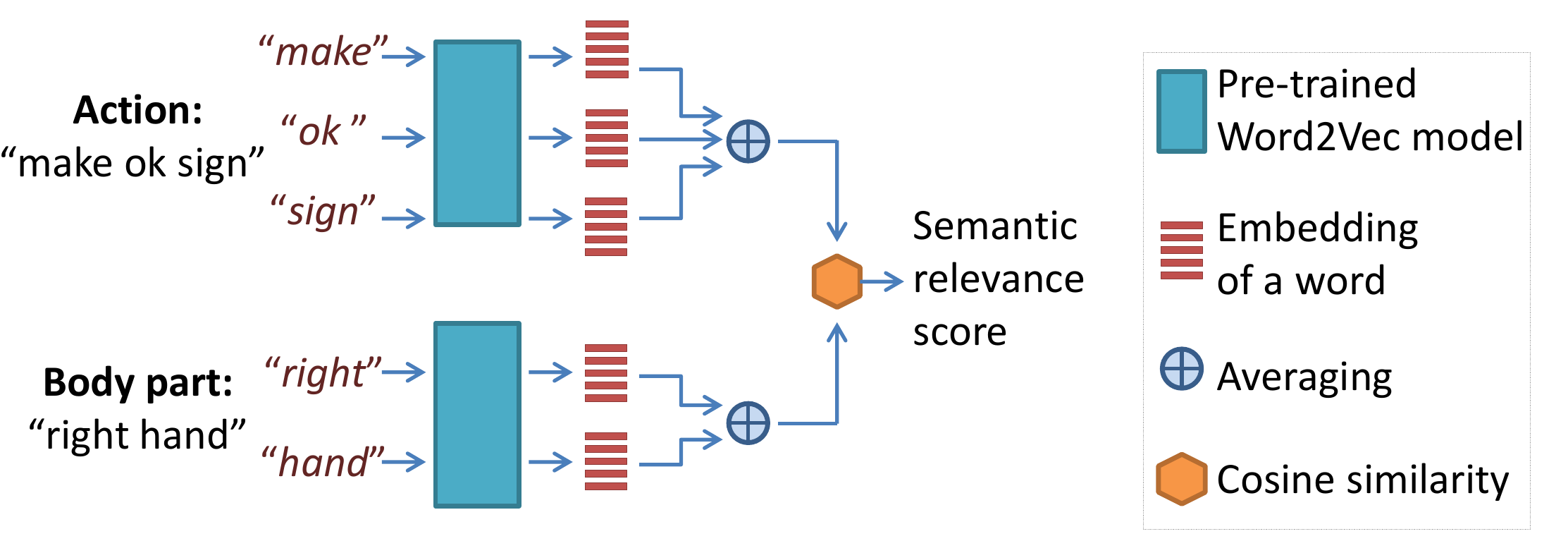}}
	\end{minipage}
\caption{Illustration of estimating the semantic relevance score between the novel action's text description (name) and the body part's text description (name).
Here we take the action ``make ok sign'' with the body part ``right hand'' as an example.
Each word in a text description is fed to the pre-trained Word2Vec model to produce its embedding (a 300-dimensional vector),
and the representation of a text description is obtained by averaging the embedding of all the words in it.
Finally, the semantic relevance is estimated by calculating the cosine similarity between the two representations.
}
\label{fig:proceduresemantic}
\end{figure}

Specifically, the powerful Word2Vec model \cite{mikolov2013distributed} that is pre-trained on a large corpus is used in our method.
When given a new action,
we estimate the relevance score (cosine similarity) between this action and each body part based on their text descriptions (\eg, ``make ok sign'' versus ``right hand''),
by using the pre-trained Word2Vec model.
For a description consisting of multiple words, its representation is obtained by averaging the embedding of all the words of it.
If the estimated relevance score is negative, we reset it to zero.
The method of estimating the semantic relevance score is illustrated in \figurename{~\ref{fig:proceduresemantic}}.

\begin{figure}
	\centering
	\includegraphics[width=0.99\linewidth,trim={0.0cm 13.0cm 6.3cm 0.0cm},clip]{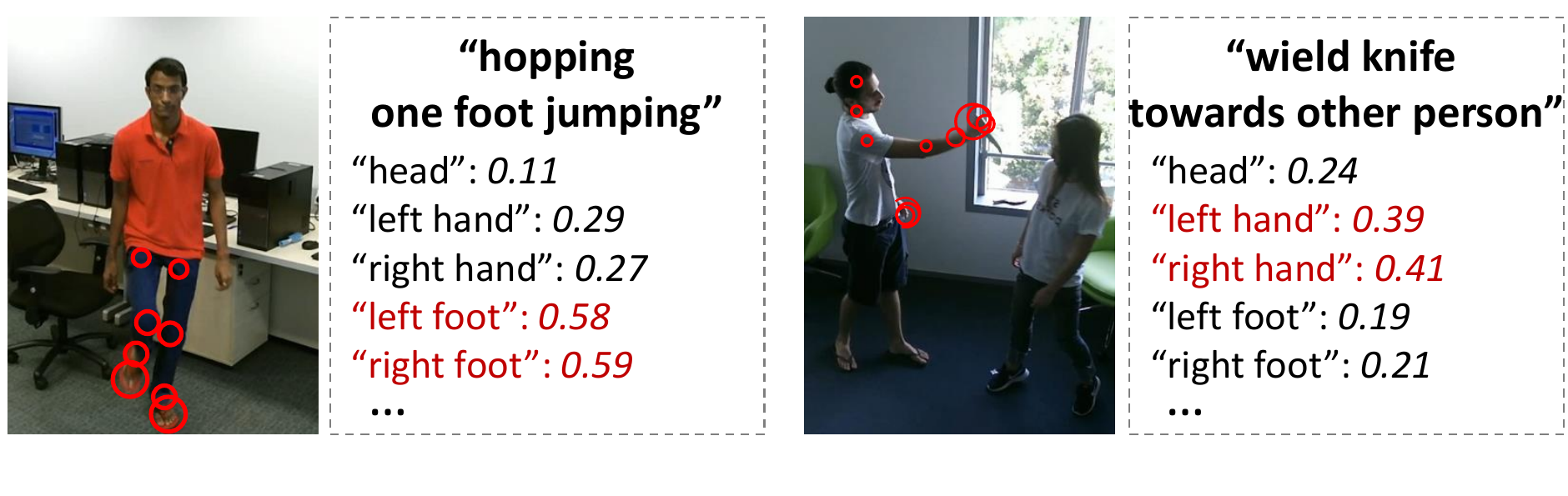}
	\caption{Examples of semantic relevance scores between action's name and each body part's name.
            In the 1st and 3rd columns, the body parts (joints) with larger scores are labeled with red circles.
            In the 2nd and 4th columns, we show the scores of several body parts.
            These scores are obtained by using pre-trained word embedding model \cite{mikolov2013distributed}.
            Semantically, ``right foot'' and ``left foot'' are very relevant to ``hopping, one foot jumping'',
            while ``right hand'' and ``left hand'' are relevant to ``wield knife towards other person''.
	}
	\label{fig:visualization}
\end{figure}

As shown in \figurename{~\ref{fig:visualization}}, the relevant body parts of the novel actions can be reliably indicated by using this method.

After we obtain the semantic relevance score ($r_{c,p}$) between the action class ($c$) and each body part ($p$),
we normalize the score as: $s_{c,p} = r_{c,p} / \sum_{u=1}^{P} r_{c,u} $
to ensure that the total score of all body parts for this action class is 1 after normalization.
Therefore, the action-part relevance score set ($\mathbf{S}_c$) of the action class ($c$) is obtained as:
$\mathbf{S}_c = \{ s_{c,p} ~ | ~ p \in \{1, ..., P\}  \}$,
which will be used as prior information for the feature generation network training and one-shot recognition evaluation.

\textbf{Training.}
In our framework,
we train the feature generation network on the auxiliary set that does not contain samples from the novel action categories.
To train this network,
at each unit $(p,t)$ (see \figurename{~\ref{fig:stlstm}}),
we feed the produced body part feature $F_{p,t}$ to the softmax classifier for action classification, similar to \cite{liu2017skeleton_PAMI}.
This indicates that at each unit, a prediction of the action class is produced based on the body part feature ($F_{p,t}$).
We train the feature generation network with the classifiers in an end-to-end manner by using the following loss function:
\begin{equation}
L = \sum_{p=1}^P \sum_{t=1}^T s_{c,p}  l(c, \hat{c}_{p,t})
\end{equation}
where $l(c, \hat{c}_{p,t})$ is the negative log-likelihood loss measuring the difference
between the true class label, $c$, of the sample and the prediction result, $\hat{c}_{p,t}$, at the unit $(p,t)$.
The semantic relevance score $s_{c,p}$ is used here as the weight of the classification loss at the units $(p,\cdot)$ that correspond to the body part $p$,
~\ie, more relevant parts are given larger loss weights.
This drives the network to learn more discriminative features on the more relevant body parts of each action class.

\textbf{Evaluation.}
After training the network on the auxiliary set,
we feed each sample from the evaluation set to the feature generation network to produce a feature set ($\mathbf{F}$) for this sample,
which represents the features of different body parts at each frame.
Note that during evaluation, we remove the classifiers and only use the produced features,
since the classes for evaluation are not contained in the auxiliary training set.

We perform weighted pooling over the obtained features ($\mathbf{F}$) of each sample to generate an aggregated representation.
Let $f (a, \mathbf{b})$ denote the aggregated representation of the sample $a$ when using $\mathbf{b}$ as the weights for feature pooling.
Then the aggregated representation of an exemplar ($\Omega$) of a novel class is calculated as:
\begin{equation}
f (\Omega, \mathbf{S}_{c_\Omega}) = \sum_{p=1}^P \sum_{t=1}^T s_{c_\Omega,p} F_{p,t}
\end{equation}
where $c_\Omega$ denotes the class label of the exemplar $\Omega$,
and $\mathbf{S}_{c_\Omega}$ is the action-part relevance score set of the action class $c_\Omega$.
Here the semantic relevance scores are used as weights
to guide the aggregated representation of the sample $\Omega$ and to emphasize the features from more relevant body parts.

To test a sample ($i$),
we calculate the distance between this sample ($i$) and an exemplar ($\Omega$) as:
\begin{equation}
D(i, \Omega) = D_{\cos} \Big( f (i, \mathbf{S}_{c_\Omega})   ,    f (\Omega, \mathbf{S}_{c_\Omega})  \Big)
\label{eq:distance}
\end{equation}
where $D_{\cos}$ is the cosine distance between the two representations (vectors).
Note that when calculating the distance of $i$ to the exemplar $\Omega$,
we generate its aggregated representation as $f (i, \mathbf{S}_{c_\Omega})$,
which indicates the weighted pooling on the feature set of $i$ is based on the relevance score set of the exemplar's class $c_\Omega$.

For each testing sample, we calculate its distances to all the exemplars from all action categories by using Eq. (\ref{eq:distance}),
and then perform classification using the nearest neighbour classifier, as in \cite{liu2017sphereface}.

\begin{table*}
    	\caption{The results of different methods, which are designed for 3D human activity analysis,
                using the cross-subject and cross-setup evaluation criteria on the NTU RGB+D 120 dataset.
        }
	\begin{center}
        \small
		\begin{tabular}{llcc}
			\toprule
            Method  &  ~~~~~~~~~~~ & Cross-Subject Accuracy  & Cross-Setup Accuracy   \\
            \toprule
			Part-Aware LSTM &\cite{Amir-Dataset-CVPR}             &  25.5\%  &  26.3\%   \\ \midrule
            Soft RNN &\cite{hu2018early}                          &  36.3\%  &  44.9\%   \\ \midrule
            Dynamic Skeleton &\cite{hu2017jointly}                &  50.8\%  &  54.7\%   \\ \midrule
            Spatio-Temporal LSTM &\cite{Liu_2016_ECCV}            &  55.7\%  &  57.9\%   \\ \midrule
            Internal Feature Fusion &\cite{liu2017skeleton_PAMI}  &  58.2\%  &  60.9\%   \\ \midrule
            GCA-LSTM &\cite{liu2017global_CVPR}                   &  58.3\%  &  59.2\%   \\ \midrule
            Multi-Task Learning Network &\cite{ke2017new_CVPR}        &  58.4\%  &  57.9\%  \\ \midrule
            FSNet &\cite{liu2019ssnet_pami}                       &  59.9\%  &  62.4\%   \\ \midrule
            Skeleton Visualization (Single Stream) &\cite{liu2017enhanced}        &  60.3\%  &  63.2\%  \\ \midrule
            Two-Stream Attention LSTM &\cite{liu2018skeleton_TIP}      &  61.2\%  &  63.3\%   \\ \midrule
            Multi-Task CNN with RotClips &\cite{ke2018learning_TIP}        &  62.2\%  &  61.8\%  \\ \midrule
            Body Pose Evolution Map &\cite{liu2018recognizing}         &  64.6\%  &  66.9\%  \\ 
            \bottomrule
		\end{tabular}
	\end{center}
	\label{tab:stateoftheart_results}
\end{table*}

\begin{table*}
    	\caption{Evaluation of using different data modalities (RGB, depth, and 3D skeleton data) for action recognition on the NTU RGB+D 120 dataset. }
	\begin{center}
        \small
		\begin{tabular}{lcc}
			\toprule
            Data Modality                                  & Cross-Subject Accuracy & Cross-Setup Accuracy    \\
            \toprule
			RGB Video                                      &  58.5\%  &  54.8\%   \\ \midrule
            Depth Video                                    &  48.7\%  &  40.1\%   \\ \midrule
            3D Skeleton Sequence                           &  55.7\%  &  57.9\%   \\ \toprule
            RGB Video + Depth Video                        &  61.9\%  &  59.2\%   \\ \midrule
            RGB Video + 3D Skeleton Sequence               &  61.2\%  &  63.1\%   \\ \midrule
            Depth Video + 3D Skeleton Sequence             &  59.2\%  &  61.2\%   \\ \midrule
            RGB Video + Depth Video + 3D Skeleton Sequence &  64.0\%  &  66.1\%   \\ 
            \bottomrule
		\end{tabular}
	\end{center}
	\label{tab:modalities_results}
\end{table*}

\section{Experiments}
\label{sec:exp}

In this section,
some state-of-the-art methods that were designed for 3D action recognition are evaluated based on the suggested cross-subject and cross-setup criteria.
Then the action recognition performances achieved by adopting different data modalities are compared.
The performance of the proposed APSR framework for one-shot 3D action recognition is also evaluated.

\subsection{Experimental Evaluations of 3D Action Recognition}
\label{sec:expeval}

\subsubsection{Evaluation of state-of-the-art methods}

Twelve state-of-the-art 3D action recognition methods are evaluated on our dataset,
namely,
the Part-Aware LSTM \cite{Amir-Dataset-CVPR},
the Soft RNN \cite{hu2018early},
the Dynamic Skeleton \cite{hu2017jointly},
the Spatio-Temporal LSTM \cite{Liu_2016_ECCV},
the Internal Feature Fusion \cite{liu2017skeleton_PAMI},
the GCA-LSTM \cite{liu2017global_CVPR},
the Multi-Task Learning Network \cite{ke2017new_CVPR},
the FSNet \cite{liu2019ssnet_pami,liu2018ssnet_CVPR},
the Skeleton Visualization \cite{liu2017enhanced},
the Two-Stream Attention LSTM \cite{liu2018skeleton_TIP},
the Multi-Task CNN with RotClips \cite{ke2018learning_TIP},
and the Body Pose Evolution Map method \cite{liu2018recognizing}.
Using the cross-subject and cross-setup evaluation criteria,
the results of these methods are reported in \tablename{~\ref{tab:stateoftheart_results}}.
%

Among these approaches, the method in \cite{hu2017jointly} uses the Fourier temporal pyramid \cite{actionletPAMI} and the hand-crafted features for action classification.
The methods in \cite{Amir-Dataset-CVPR,hu2018early,Liu_2016_ECCV,liu2017skeleton_PAMI,liu2017global_CVPR,liu2018skeleton_TIP} all use RNN/LSTM for 3D action recognition,
and the approaches in \cite{ke2017new_CVPR,liu2019ssnet_pami,liu2017enhanced,ke2018learning_TIP,liu2018recognizing} all use convolutional networks for spatial and temporal modeling.
Specifically, the evaluations of \cite{liu2017enhanced,liu2018recognizing} are performed
by using the efficient MobileNet \cite{howard2017mobilenets} as the base model, and skeleton data as input.
All the evaluated implementations are from the original authors of the corresponding papers.

\subsubsection{Evaluation of using different data modalities}

Since multiple data modalities are provided in our dataset,
we also evaluate the performance of using different data modalities (\eg, RGB, depth, and skeleton data) as input for action recognition,
and report the results in \tablename{~\ref{tab:modalities_results}}.

In this table, the accuracy on RGB video is achieved by learning visual and motion (optical flow) features at each frame of the RGB video by training the VGG model \cite{simonyan2014very},
and performing classification by averaging the classification results of all frames, similarly to \cite{twostreamCNN}.
The accuracy on depth video is obtained by using the similar method as the RGB video, and training the VGG model based on the depth data.
The accuracy on 3D skeleton data is achieved by using the method in \cite{Liu_2016_ECCV}.

We observe that when using the RGB or depth video as input,
the performance of the cross-setup evaluation is weaker than that of the cross-subject one.
The performance disparity can be justified as:
(1) In the cross-setup evaluation scenario, the heights and distances of the cameras are not the same over different setups.
This indicates camera viewpoints are different, and thus the appearance of the actions can be significantly different.
(2) The background is also changing across different setups.
However, when using the RGB or depth frames as input, the methods are more prone to learn from the appearance or view-dependent motion patterns.

By using the 3D skeleton data as input, the method performs better in the cross-setup evaluation.
This is partially because the method using 3D skeleton data is stronger to generalize among different views,
since the 3D skeletal representation is view-invariant in essence.
However, it is still prone to errors of the body tracker.

We also evaluate the performance of fusing multiple data modalities for action recognition.
The results in \tablename{~\ref{tab:modalities_results}} show that compared to the method of using a single modality,
fusing multiple modalities bring a performance improvement,
since they contain complementary and discriminative appearance and geometrical information, which is useful for activity analysis.

\subsubsection{Evaluation of using different sizes of training set}

In the aforementioned experiments, the evaluated deep learning models,
such as Part-Aware LSTM \cite{Amir-Dataset-CVPR} and FSNet \cite{liu2019ssnet_pami},
are trained with the defined large training set from our NTU RGB+D 120 dataset.
Here we call this training set as ``full training set''.

To evaluate the effect of the training set size on the recognition performance,
we use different ratios of training samples from the full training set for network training.
We then evaluate the action recognition performance on the same testing set based on the cross-setup evaluation criterion.

\begin{figure}[tb]
    \begin{minipage}[h]{1.0\linewidth}
		\centering
		\centerline{\includegraphics[scale=0.6,trim={3.8cm 9.5cm 4.0cm 10.5cm},clip]{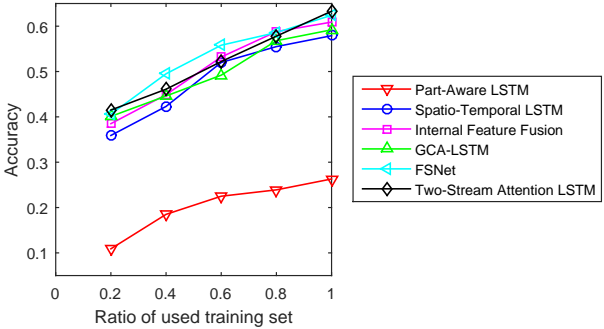}}
	\end{minipage}
\caption{Evaluation of using different sizes of training set for action recognition with different methods.
In this figure, ratio $1.0$ means the full training set is used for network training.}
\label{fig:lessdataMethod}
\end{figure}

We take six methods \cite{Amir-Dataset-CVPR,Liu_2016_ECCV,liu2017skeleton_PAMI,liu2017global_CVPR,liu2019ssnet_pami,liu2018skeleton_TIP} as examples,
and show their results achieved by using different sizes of training set in \figurename{~\ref{fig:lessdataMethod}}.
The results show that when more samples are used for network training,
the action recognition accuracies of all these methods increase obviously.
For example, when using 20\% of the samples from the full training set for training,
the accuracy of FSNet is 40.6\%,
while the accuracy reaches 62.4\% when the full training set is used for network training.

\begin{figure}[tb]
    \begin{minipage}[h]{1.0\linewidth}
		\centering
		\centerline{\includegraphics[scale=0.6,trim={3.8cm 9.5cm 4.0cm 10.5cm},clip]{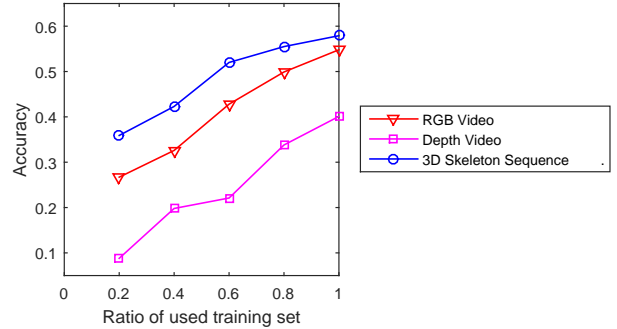}}
	\end{minipage}
\caption{Evaluation of using different sizes of training set for action recognition with different data modalities.
}
\label{fig:lessdataModality}
\end{figure}

We also evaluate the performance of using different sizes of training data for action recognition with different data modalities,
and show the results in \figurename{~\ref{fig:lessdataModality}}.
The results in this figure also show the benefit of using more data for network training to achieve better action recognition performance.

\subsubsection{Detailed analysis according to data modalities}

\begin{figure}[tbp]
    \begin{minipage}[h]{1.0\linewidth}
		\centering
		\centerline{\includegraphics[scale=0.46]{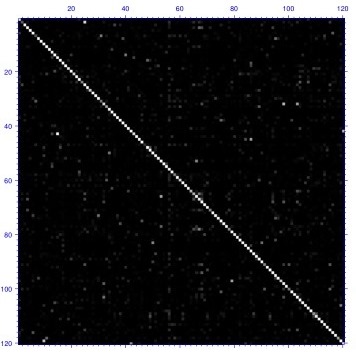}}
	\end{minipage}
\caption{Confusion matrix of the RGB data modality.}
\label{fig:confuse_matrix_RGBDS}
\end{figure}

\begin{table*}[tbp]
    \caption{Action recognition results of different data modalities on the NTU RGB+D 120 dataset.
    }
	\begin{center}
        \scriptsize
		\begin{tabular}{cll}
\toprule
~~~~~~~~~~\textbf{Data Modality}~~~~~~~~~~         &  \textbf{Top 10 accurate actions}~~~~~~~~~~~~~~~~~~~~~~~~~~~~~~         &  \textbf{Top 10 confused (misclassified) action pairs}  \\
\toprule
&	1. walk apart from each other	&	1. take off a shoe$\rightarrow$put on a shoe \\
&	2. walk towards each other	&	2. kick backward$\rightarrow$side kick \\
&	3. hopping	&	3. rub two hands together$\rightarrow$clapping\\
&	4. carry things with other person	&	4. reading$\rightarrow$writing \\
RGB &	5. arm swings	&	5. clapping$\rightarrow$rub two hands together \\
&	6. staggering	&	6. vomiting condition$\rightarrow$bow \\
&	7. pick up things	&	7. ball up paper$\rightarrow$fold paper \\
&	8. put on jacket	&	8. open a box$\rightarrow$fold paper \\
&	9. hugging other person	&	9. both hands up$\rightarrow$cheer up \\
&	10. move heavy objects	&	10. yawn$\rightarrow$blow nose\\
\midrule			
&	1. carry things with other person	&	1. take off a shoe$\rightarrow$put on a shoe \\
&	2. walk apart from each other	&	2. reading$\rightarrow$writing \\
&	3. move heavy objects	&	3. playing with tablet$\rightarrow$writing \\
&	4. hugging other person	&	4. bow$\rightarrow$vomiting condition \\
Depth &	5. kick backward	&	5. both hands up$\rightarrow$stretch oneself \\
&	6. walk towards each other	&	6. put on jacket$\rightarrow$put on bag/backpack \\
&	7. hopping	&	7. vomiting condition$\rightarrow$bow \\
&	8. arm swings	&	8. cheer up$\rightarrow$both hands up \\
&	9. staggering	&	9. rub two hands together$\rightarrow$clapping \\
&	10. open a box	&	10. take off bag/backpack$\rightarrow$take off jacket \\
\midrule			
&	1. walk apart from each other	&	1. put on a shoe$\rightarrow$take off a shoe \\
&	2. standing up	&	2. hit other person with object$\rightarrow$wield knife towards other person \\
&	3. walk towards each other	&	3. make ok sign$\rightarrow$make victory sign \\
&	4. hugging other person	&	4. thumb up$\rightarrow$make victory sign \\
Skeleton &	5. arm swings	&	5. put on jacket$\rightarrow$put on bag/backpack \\
&	6. squat down	&	6. touch other person's pocket (steal)$\rightarrow$grab other person's stuff \\ 
&	7. sitting down	&	7. make victory sign$\rightarrow$make ok sign \\
&	8. pushing other person	&	8. play magic cube$\rightarrow$counting money \\
&	9. arm circles	&	9. take a photo of other person$\rightarrow$shoot at other person with a gun \\
&	10. kick backward	&	10. handshaking$\rightarrow$giving something to other person \\
\toprule			
&	1. walk apart from each other	&	1. take off a shoe$\rightarrow$put on a shoe \\
&	2. carry things with other person	&	2. vomiting condition$\rightarrow$bow \\
&	3. hugging other person	&	3. both hands up$\rightarrow$stretch oneself \\
RGB &	4. walk towards each other	&	4. clapping$\rightarrow$rub two hands together \\
+ Depth&	5. standing up  &	5. yawn$\rightarrow$blow nose \\
+ Skeleton& 6. hopping	&	6. put on a shoe$\rightarrow$take off a shoe \\
&	7. squat down &	7. hush (say quite)$\rightarrow$blow nose \\
&	8. move heavy objects &	8. rub two hands together$\rightarrow$clapping \\
&	9. arm swings &	9. reading$\rightarrow$writing \\
&	10. put on jacket &	10. stretch oneself$\rightarrow$both hands up \\
            \bottomrule
		\end{tabular}
	\end{center}
(1) Top 10 accurate actions denote the actions that have the top 10 recognition accuracies.\\
(2) Top 10 confused (misclassified) action pairs denote the action pairs that have the top 10 confusion rates (misclassification percentages).\\
(3) $A$$\rightarrow$$B$ denotes a confused action pair, where some samples of class $A$ are misclassified to class $B$.\\
\label{tab:ActionWiseRGBDS}
\end{table*}


Here we analyze the results obtained by using different data modalities in detail.
The skeleton data modality is evaluated using the Spatio-Temporal LSTM \cite{Liu_2016_ECCV}.
The RGB and depth data modalities are both evaluated using the two-stream framework \cite{twostreamCNN}.
Modality fusion is performed by fusing the results of the three modalities.


We first plot the confusion matrices of different data modalities.
Specifically, we show the confusion matrix of the RGB modality as an example in \figurename{~\ref{fig:confuse_matrix_RGBDS}}.

We then perform action-wise analysis for different data modalities. 
Considering the large number of action categories,
for each data modality, we analyze the action classes that have high recognition accuracies (top 10 accurate classes),
and the actions that are easily misclassified to other classes (top 10 confused action pairs).
We show the results in \tablename{~\ref{tab:ActionWiseRGBDS}}.


Based on the results in \tablename{~\ref{tab:ActionWiseRGBDS}},
we find that the actions that have significant motions and discriminative posture patterns could be more accurately recognized.
For example, the actions ``walk apart from each other'' and ``walk towards each other'', which have very discriminative and significant motions,
are both in the top 10 accurate actions when using any of the three data modalities as input.

We also observe that when using skeleton data as input,
the actions that involve interactions with objects may be easily misclassified.
For example, ``play magic cube'' is often confused with ``counting money'',
and ``handshaking'' can also be misclassified to ``giving something to other person'',
as shown in \tablename{~\ref{tab:ActionWiseRGBDS}}.
This is possibly because the actions in each pair have similar human motion patterns,
and the perception of the existences of the objects and their appearances is important for accurately recognizing these actions.
However, the appearance information of the objects is ignored by the recognition model when using skeleton information only.
In contrast, when we perform action recognition based on the RGB or depth data that captures the object information,
many object-related actions could be accurately recognized.
For example, ``carry things with other person'' and ``move heavy objects'' are both in the top 10 accurate actions of the RGB and depth data modalities.

Although RGB and depth data modalities both have good ability in representing the object-related actions,
the actions with object involved may be misclassified when the same (or similar) objects are shared by different actions,
where the objects may even mislead the classification.
For example, when using RGB data as input, ``ball up paper'' tend to be misclassified to ``fold paper'',
and ``reading'' and ``writing'' can also be confused, as shown in \tablename{~\ref{tab:ActionWiseRGBDS}}.
An interesting observation is that for RGB data modality,
many samples of ``open a box'' are misclassified to ``fold paper'',
while ``open a box'' is in the top 10 accurate actions for the depth data modality.
This performance disparity could be explained as: in the RGB image, the appearances of the box and the paper can be similar,
but the depth data can well represent the 3D shape information of the objects,
thus depth data is more powerful in distinguishing the box from paper.
As a result, the action ``open a box'' can be more accurately recognized by using depth data than by using RGB data.
We also observe that ``put on jacket'', which is classified well with RGB data, is easily confused with ``put on bag/backpack'' with depth data.
This may be because the jacket and the backpack can be more easily distinguished from their color and texture information, than from their 3D shape information.

As shown in \tablename{~\ref{tab:ActionWiseRGBDS}}, ``kick backward'' is in the top 10 accurate actions of both skeleton and depth data modalities,
while ``kick backward'' is easily confused to ``side kick'' when using RGB data.
A possible reason is that both the depth data modality and the skeleton data modality
provide the 3D structure information that implies the 3D direction information of the body part motions.
However, such 3D information is not provided when using the RGB data.

We also observe that many actions that contain fine-grained hand gestures and finger motions,
such as ``make ok sign'', ``make victory sign'', and ``thumb up'', are easily misclassified when using the skeleton data only, as depicted in \tablename{~\ref{tab:ActionWiseRGBDS}}.
The performance limitation of the skeleton data in handing these actions
is possibly due to that only three joints are provided for each hand in the skeleton data,
and besides, 
the skeleton data provided by Kinect's tracker algorithm is not perfect and can be noisy sometimes.
This indicates that the skeleton data has difficulties in representing the very detailed hand and finger motions.
Therefore, the actions with fine-grained hand/finger motions could be easily misclassified when using the skeleton data only.

In \tablename{~\ref{tab:ActionWiseRGBDS}}, we also find that there are several \emph{tough} action pairs that are easily confused for all the data modalities,
such as ``take off a shoe'' and ``put on a shoe''.
A possible explantation is that the human motion and object appearance information in these actions are both very similar, and thus they are difficult to be accurately distinguished.

In summary, the 2D appearance information (\emph{e.g.}, color and texture) and the 3D shape and structure information provided by different data modalities
can all affect the recognition performance of different types of actions.

In \tablename{~\ref{tab:ActionWiseRGBDS}},
beside presenting the top 10 accurate actions and the top 10 confused action pairs for different data modalities,
we also show the results of fusing the three modalities.
We observe that when fusing these modalities which provide complementary appearance and geometric information,
the recognition performance is improved.
For example, when using RGB, depth, or skeleton data only,
the recognition accuracies of ``walk apart from each other'' are 94\%, 93\%, and 92\%, respectively.
When fusing the three modalities, the accuracy reaches 99\%.
We also find that the confusion rates of the \emph{tough} action pairs can also drop when fusing the three data modalities.
For the very \emph{tough} action pair: ``take off a shoe'' and ``put on a shoe'',
the confusion rates for different single modalities are 52\%, 65\%, and 39\%, respectively,
and the confusion rate decreases to 32\% with modality fusion.

\subsubsection{Detailed analysis according to methods}

We also analyze the experimental results of different 3D action recognition methods on our dataset in detail.
We take five state-of-the-art methods as examples for analysis, namely
Internal Feature Fusion \cite{liu2017skeleton_PAMI},
GCA-LSTM \cite{liu2017global_CVPR},
Multi-Task Learning Network \cite{ke2017new_CVPR},
FSNet \cite{liu2019ssnet_pami},
and Multi-Task CNN with RotClips \cite{ke2018learning_TIP}.

\begin{figure}[tbp]
    \begin{minipage}[h]{1.0\linewidth}
		\centering
		\centerline{\includegraphics[scale=0.46]{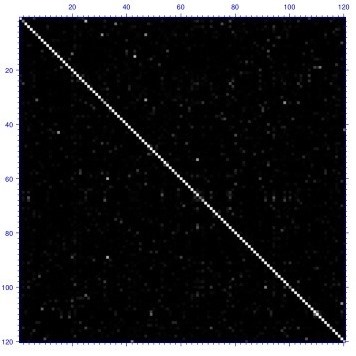}}
	\end{minipage}
\caption{Confusion matrix of Internal Feature Fusion \cite{liu2017skeleton_PAMI}.}
\label{fig:confuse_matrix_methods}
\end{figure}

We first plot the confusion matrices of these methods.
The confusion matrix of the method, Internal Feature Fusion, is shown in \figurename{~\ref{fig:confuse_matrix_methods}} as an example.

We then perform action-wise analysis for these methods.
Specifically, we perform detailed analysis for the top 10 accurate actions
and the top 10 confused action pairs of each method (see \tablename{~\ref{tab:ActionWiseMethods}}),
considering the large number of action classes.

\begin{table*}[tbp]
    \caption{Action recognition results of different methods on the NTU RGB+D 120 dataset.}
	\begin{center}
        \scriptsize
		\begin{tabular}{cll}
\toprule
~~~~~~~~~~~~~~\textbf{Method}~~~~~~~~~~~~~~         &  \textbf{Top 10 accurate actions}~~~~~~~~~~~~~~~~~~~~~~~~~~~~~~         &  \textbf{Top 10 confused (misclassified) action pairs}  \\
\toprule					
&	1.	walk apart from each other	&	1.	put on a shoe$\rightarrow$take off a shoe	\\
&	2.	walk towards each other	&	2.	blow nose$\rightarrow$hush (say quite)	\\
&	3.	carry things with other person	&	3.	clapping$\rightarrow$rub two hands together	\\
Internal &	4.	hugging other person	&	4.	take off a shoe$\rightarrow$put on a shoe	\\
Feature &	5.	standing up	&	5.	rub two hands together$\rightarrow$clapping	\\
Fusion &	6.	kick backward	&	6.	stretch oneself$\rightarrow$both hands up \\
\cite{liu2017skeleton_PAMI} &	7.	take off jacket &	7.	yawn$\rightarrow$blow nose	\\
&	8.	arm swings &	8.	bow$\rightarrow$vomiting condition	\\
&	9.	put on jacket &	9.	grab other person's stuff$\rightarrow$touch other person's pocket (steal) \\
&	10.	bounce ball &	10.	vomiting condition$\rightarrow$sneeze/cough	\\
\midrule						
&	1.	walk apart from each other	&	1.	put on a shoe$\rightarrow$take off a shoe	\\
&	2.	standing up	&	2.	play magic cube$\rightarrow$counting money	\\
&	3.	walk towards each other	&	3.	make victory sign$\rightarrow$make ok sign	\\
&	4.	hugging other person	&	4.	hit other person with object$\rightarrow$wield knife towards other person	\\
GCA-LSTM &	5.	high five	&	5.	take something out of a bag/backpack$\rightarrow$put something into a bag/backpack	\\
\cite{liu2017global_CVPR} &	6.	handshaking	&	6.	kick backward$\rightarrow$hopping \\
&	7.	arm swings	&	7.	staggering$\rightarrow$kick backward \\
&	8.	sitting down	&	8.	put on jacket$\rightarrow$put on bag/backpack \\ 
&	9.	arm circles	&	9.	giving something to other person$\rightarrow$exchange things with other person	\\
&	10.	squat down	&	10.	grab other person's stuff$\rightarrow$touch other person's pocket (steal) \\
\midrule						
&	1.	cross toe touch	&	1.	giving something to other person$\rightarrow$exchange things with other person	\\
&	2.	walk apart from each other	&	2.	point finger at other person$\rightarrow$shoot at other person with a gun	\\
&	3.	walk towards each other	&	3.	slapping other person$\rightarrow$hit other person with object	\\
Multi-Task &	4.	arm swings	&	4.	rub two hands together$\rightarrow$apply cream on hand back	\\
Learning &	5.	kick backward	&	5.	cutting paper using scissors$\rightarrow$staple book	\\
 Network &	6.	cheers and drink	&	6.	yawn$\rightarrow$hush (say quite) \\
\cite{ke2017new_CVPR} &	7.	squat down	&	7.	take off glasses$\rightarrow$take off headphone	\\
&	8.	arm circles	&	8.	make ok sign$\rightarrow$thumb up	\\
&	9.	running on the spot	&	9.	take off a shoe$\rightarrow$put on a shoe	\\
&	10.	cheer up	&	10.	make ok sign$\rightarrow$make victory sign	\\
\midrule						
&	1.	standing up	&	1.	put on a shoe$\rightarrow$take off a shoe	\\
&	2.	arm circles	&	2.	make victory sign$\rightarrow$make ok sign	\\
&	3.	walk apart from each other	&	3.	put on jacket$\rightarrow$put on bag/backpack \\ 
&	4.	kick backward	&	4.	hit other person with object$\rightarrow$wield knife towards other person	\\
FSNet &	5.	arm swings	&	5.	counting money$\rightarrow$play magic cube	\\
\cite{liu2019ssnet_pami} &	6.	cross toe touch	&	6.	rub two hands together$\rightarrow$clapping	\\
&	7.	grab other person's stuff	&	7.	pushing other person$\rightarrow$slapping other person	\\
&	8.	cheer up	&	8.	pat on back of other person$\rightarrow$hit other person with object \\ 
&	9.	running on the spot	&	9.	kicking other person$\rightarrow$step on foot of other person	\\
&	10.	walk towards each other &	10.	hit other person with body$\rightarrow$support other person with hand \\
\midrule						
&	1.	cross toe touch	&	1.	slapping other person$\rightarrow$hit other person with object	\\
&	2.	walk apart from each other	&	2.	point finger at other person$\rightarrow$shoot at other person with a gun	\\
&	3.	running on the spot	&	3.	make victory sign$\rightarrow$make ok sign	\\
Multi-Task &	4.	arm circles	&	4.	tear up paper$\rightarrow$open a box	\\
CNN with &	5.	walk towards each other	&	5.	staple book$\rightarrow$cutting paper using scissors	\\
RotClips &	6.	squat down	&	6.	yawn$\rightarrow$hush (say quite)	\\
\cite{ke2018learning_TIP} &	7.	arm swings	&	7.	giving something to other person$\rightarrow$exchange things with other person	\\
&	8.	kick backward	&	8.	playing with tablet$\rightarrow$play magic cube	\\
&	9.	cheer up	&	9.	take off glasses$\rightarrow$take off headphone \\
&	10.	playing rock-paper-scissors	&	10.	take off a shoe$\rightarrow$put on a shoe	\\
\bottomrule
		\end{tabular}
	\end{center}
	\label{tab:ActionWiseMethods}
\end{table*}


Among these methods, 
the method Internal Feature Fusion performs action recognition by fusing the 3D skeleton-based geometric features and the RGB-based appearance features,
and the other four approaches all use 3D skeleton data as input for 3D action recognition.

In \tablename{~\ref{tab:ActionWiseMethods}}, we observe that the top 10 confused action pairs of GCA-LSTM, Multi-Task Learning Network, FSNet, and Multi-Task CNN with RotClips
all contain many object-related actions (such as ``put on jacket'' and ``play magic cube'')
and fine-grained hand/finger motion-based actions (such as ``make victory sign'' and ``thumb up'').
This is possibly owing to that all of these four approaches perform action recognition based on the 3D skeleton data
that is not able to represent the object information and the fine-grained finger motions well.
Therefore, these approaches have difficulties in dealing with the object-related and fine-grained hand/finger motion-based activities.
In contrast, we observe that there are many object-related actions (such as ``put on jacket'', ``bounce ball'', and ``carry things with other person'')
in the top 10 accurate actions of the method Internal Feature Fusion, which uses both 3D skeleton data and RGB data as input for action recognition.

In this table, there are also some actions (\emph{e.g.}, ``take off a shoe'' and ``put on a shoe'') that are very similar in motions and appearances,
and are difficult to be distinguished well by all the five methods.

We also observe that the actions ``walk apart from each other'' and ``walk towards each other''
that have significant motions and discriminative posture patterns are in the top 10 accurate actions of all the methods.

In \tablename{~\ref{tab:ActionWiseMethods}}, we observe the top 10 confused action pairs of FSNet contain some two-person mutual action pairs
without object involved, 
such as ``hit other person with body'' with ``support other person with hand'', and ``kicking other person'' with ``step on foot of other person'',
while these actions can be classified relatively reliably by the other approaches.
A possible explanation of the performance limitation of FSNet in handling these mutual actions is that
in FSNet, the features of the two persons are extracted separately, and these features are then simply averaged for action recognition.
This indicates the interaction patterns between the two persons are not well represented,
and thus the mutual actions may be easily misclassified by FSNet.

We also observe that the action ``kick backward'' is easily confused with ``hopping'' by the method GCA-LSTM,
while ``kick backward'' is in the top 10 accurate actions of all the other four methods.
A possible reason is that GCA-LSTM normalizes the skeleton data for the single-person actions
by rotating the skeleton to the frontal view and translating the body center to the origin in each frame at the pre-processing stage.
After such normalization, ``hopping'' could be similar to ``kick backward'',
since the vertical movements of the body center in ``hopping'' are ignored by the method GCA-LSTM.

In \tablename{~\ref{tab:ActionWiseMethods}}, ``grab other person's stuff'' is in the top 10 accurate actions of the method FSNet,
while ``grab other person's stuff'' can be easily confused with ``touch other person's pocket (steal)'' by Internal Feature Fusion and GCA-LSTM.
A possible explanation of the performance disparity is that FSNet may be able to learn the motion speed information better than the other two methods.
The human postures of these two actions are quite similar and their main difference is the motion speed,
\emph{i.e.}, the motions in ``touch other person's pocket (steal)'' are very slow,
while the motions in ``grab other person's stuff'' are much faster.
Both Internal Feature Fusion and GCA-LSTM use recurrent models to learn the temporal dynamics of the actions based on the sampled 20 frames from each action sequence.
This implies the speed information of the actions may be ignored by them.
In contrast, FSNet uses a temporal convolutional model to learn the temporal context information over all the frames of each action sample,
and thus is able to better learn the motion speed information, which is an important cue to distinguish ``grab other person's stuff'' from ``touch other person's pocket (steal)''.

\subsection{Experimental Evaluations of One-Shot Recognition}
\label{sec:expevaloneshot}

We evaluate the one-shot recognition performance on our dataset.
In our experiments, the auxiliary set for feature generation network training contains 100 action classes,
and the evaluation set for one-shot recognition evaluation contains the remaining 20 classes.

We compare the following methods for one-shot 3D action recognition:

(1) Average Pooling.
This method is similar to the approach in \cite{liu2017skeleton_PAMI}.
To adapt \cite{liu2017skeleton_PAMI} for one-shot recognition,
we use ST-LSTM \cite{liu2017skeleton_PAMI} as the feature generation network to learn the features of the body parts at each frame,
and during evaluation, the features of all body parts at all frames are aggregated with average pooling.
The distance between the testing sample and each exemplar is calculated using the average pooling representation of each video.

(2) Fully Connected.
In this method, the feature generation network is constructed by adding a fully connected layer above the ST-LSTM model.
Concretely, the outputs from all spatio-temporal units of the ST-LSTM are concatenated
and fed to a fully connected layer to generate a global representation for the input video \cite{liu2017global_CVPR}.
During training, the ST-LSTM and the fully connected layer is trained in an end-to-end fashion.
During evaluation, the distance is calculated using the global representation of each video.

(3) Attention Network.
This method is similar to the above ''Fully Connected'' method,
except that an attention mechanism \cite{liu2017global_CVPR} is added to the feature generation network,
\ie, the attention scores of different joints are automatically learned in this method.

(4) APSR.
This is the proposed Action-Part Semantic Relevance-aware (APSR) framework,
which assigns different scores to different joints (\ie, weighted pooling),
by considering the sematic relevance between the novel action's name and body part's name for one-shot recognition.

\begin{table}[tb]
    	\caption{The results of one-shot 3D action recognition on the NTU RGB+D 120 dataset.}
	\begin{center}
        \small
		\begin{tabular}{lc}
			\toprule
            Method                                           &  Evaluation Accuracy \\
            \toprule
			Average Pooling \cite{liu2017skeleton_PAMI}      &  42.9\%                    \\ \midrule
            Fully Connected \cite{liu2017global_CVPR}      &  42.1\%                    \\ \midrule
            Attention Network \cite{liu2017global_CVPR}     &  41.0\%                   \\ \midrule
            APSR                                             &  \textbf{45.3}\%                   \\ 
            \bottomrule
		\end{tabular}
	\end{center}
	\label{tab:oneshot_results}
\end{table}

\begin{table}[tb]
    	\caption{The results of using different sizes of auxiliary training set to learn the feature generation network, for one-shot recognition on the novel action classes.}
	\begin{center}
        \small
		\begin{tabular}{ccc}
			\toprule
            \multicolumn{2}{c}{Auxiliary Training Set} &  ~~~Evaluation~~~ \\ \cline{1-2}
			\#Training Samples  & \#Training Classes   &  Accuracy      \\
            \toprule
             $19,000$         &    $20$               &     29.1\%     \\ \midrule
             $38,000$         &    $40$               &     34.8\%     \\ \midrule
             $57,000$         &    $60$               &     39.2\%     \\ \midrule
             $76,000$         &    $80$               &     42.8\%     \\ \midrule
			 $95,000$         &   $100$               &     45.3\%     \\ 
            \bottomrule
		\end{tabular}
	\end{center}
	\label{tab:ontshot2_results2}
\end{table}

The comparison results of these approaches are shown in \tablename{~\ref{tab:oneshot_results}}.
The proposed APSR framework achieves the best results,
which indicates the generalization capability of the proposed method on the novel action categories.
We also observe that the performance of ``Attention Network'' \cite{liu2017global_CVPR},
which learns to assign weights to different joints with the attention mechanism, is even weaker than that of ``Average Pooling'' \cite{liu2017skeleton_PAMI}.
A possible explanation is that the attention ability is trained on the auxiliary set that does not contain the novel actions,
thus when handling the novel actions, its performance is even worse than directly performing average pooling.
This further demonstrates the superiority of the introduced APSR framework.

In the aforementioned experiments, the feature generation network of our APSR framework is trained on a large auxiliary set containing about 100 thousand videos.
We also try downsizing the auxiliary training set,
and evaluate the one-shot 3D action recognition performance on the same evaluation set.
The results in \tablename{~\ref{tab:ontshot2_results2}} show that the one-shot recognition accuracy drops
and the generalization capability to novel classes is weakened, when using fewer classes and samples for learning the feature generation network.
This also implies the demand for a large dataset, and it is in line with our motivation for proposing the NTU RGB+D 120 dataset.

\subsection{Discussions}
\label{sec:discussions}

The introduction of this very large-scale and challenging dataset with high variability in different aspects
(\emph{e.g.}, subjects, environments, camera views, and action categories)
will facilitate the users to apply, adapt, develop, and evaluate various learning-based techniques for the future research on human activity analysis.
Below we discuss some of the potential research problems and techniques that could be investigated by taking advantage of our dataset:

\emph{(1) Activity analysis with different data modalities.}
Four different data modalities are provided by our dataset,
namely, depth videos, 3D skeleton data, RGB videos, and infrared sequences.
Different modalities have different structures of the data, and have different application advantages.
Therefore, users can utilize our dataset to investigate the algorithms for
depth-based, skeleton-based, RGB-based, or infrared-based action recognition.

\emph{(2) Heterogeneous feature fusion analysis.}
The provided data modalities contain complementary appearance and 3D geometrical information for human activity analysis.
Thus users can take advantage of our dataset to identify the strengths of respective modalities,
and further investigate various fusion techniques for the heterogeneous features \cite{DSSCA-PAMI}
extracted from different data modalities.

\emph{(3) Deep network pre-training.}
Most of the existing datasets for RGB+D action recognition are relative small,
thus the deep models evaluated on them often suffer from over-fitting issues.
Since the proposed dataset has a large number of samples with diversity in various factors,
it can also be employed for network pre-training.
By initializing the network parameters on our proposed large-scale dataset,
the deep models are expected to be able to generalize better on other relative small datasets for 3D activity analysis,
as analyzed in \cite{baradel2017pose}.

\emph{(4) Cross-subject activity analysis.}
In the proposed dataset, the 106 human subjects are in a wide range of age and height distribution,
and are from different cultural backgrounds.
These factors bring realistic variation to the quality of actions,
and make it possible to have more sensible cross-subject evaluations for the 3D activity analysis methods.
This also encourages the community to develop action recognition algorithms that are robust for different subjects.

\emph{(5) Cross-environment activity analysis.}
Our dataset is collected under different environmental conditions that use 96 different backgrounds with significant illumination variation.
This enables the users to perform cross-environment activity analysis.
The different collection environments can also facilitate the analysis of the algorithms' robustness against the variation in backgrounds and illuminations.

\emph{(6) Cross-view activity analysis.}
The proposed dataset is collected with 155 camera views,
which facilitates the cross-view activity analysis
and encourages the users to develop action recognition approaches that are robust against view variation for the practical applications.

\emph{(7) Cross-modal transfer learning.}
Learning representations from a large labeled modality for transfer learning for the smaller-scale new modalities has attracted a lot of research attention and has been applied to different tasks recently \cite{gupta2016cross}.
The proposed large dataset that provides different data modalities could be utilized for the research on cross-modal transfer learning.

\emph{(8) Mutual activity analysis.}
Human-human interaction analysis is also an important branch of human activity analysis.
The proposed dataset contains 25 thousand two-person mutual action videos that correspond to 26 different two-person interaction classes.
This facilitates the users to investigate and develop various approaches for handling the task of mutual action recognition.

\emph{(9) Real-time skeleton-based early action recognition.}
3D skeleton data has shown its advantages in real-time early action recognition due to its succinct and high level representation,
as analyzed in \cite{keglobal}.
This indicates our large dataset can also be used for the research on real-time early action recognition.

\emph{(10) One-shot 3D activity analysis.}
Our large-scale dataset can also be used to learn a discriminative representation model for one-shot 3D activity analysis for novel action classes.


\section{Conclusion}
\label{sec:conclusion}

A large-scale RGB+D action recognition dataset is introduced in this paper.
Our dataset includes 114,480 video samples collected from 120 action classes in highly variant camera settings.
Compared to the current datasets for this task, our dataset is larger in orders and contains much more variety in different aspects.
The large scale of the collected data facilitates us to apply data-driven learning methods to this problem and achieve promising performance.
We also propose an APSR framework for one-shot 3D action recognition.
The provided experimental results show the availability of large-scale data enables the data-driven learning frameworks to achieve promising results.

\begin{figure*}
	\centering
	\setlength{\tabcolsep}{2pt} 
	\begin{tabular}{ccccc}		
		\includegraphics[width=85pt]{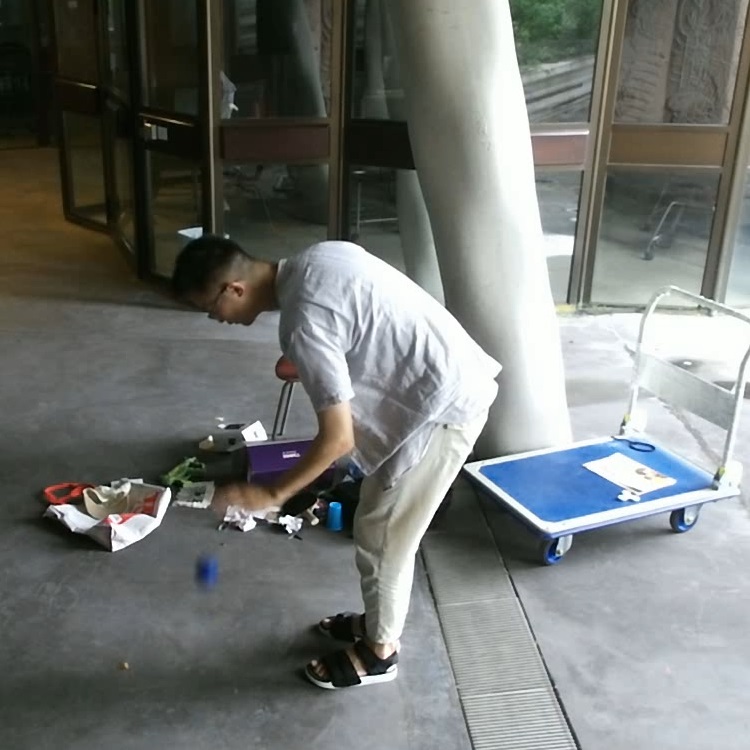} &
		\includegraphics[width=85pt]{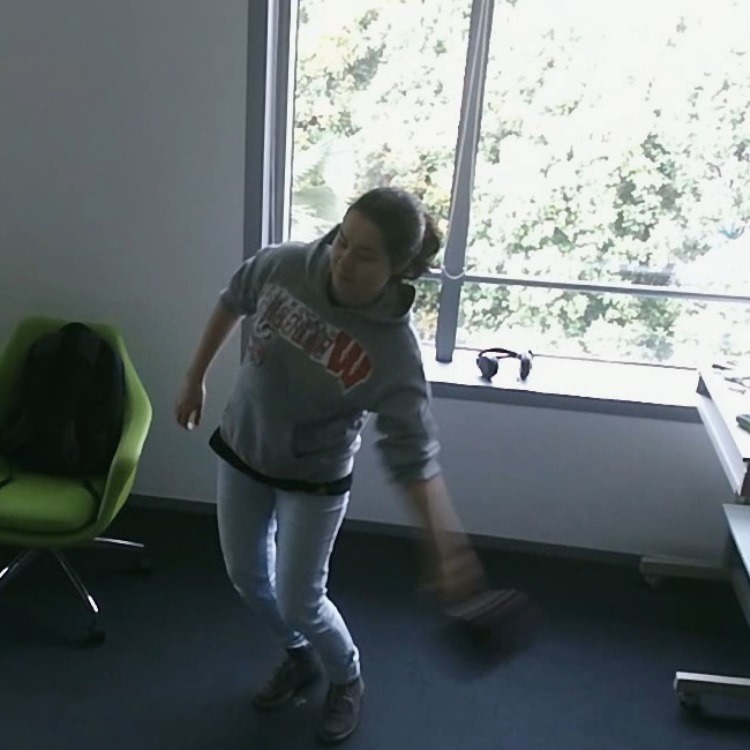} &
		\includegraphics[width=85pt]{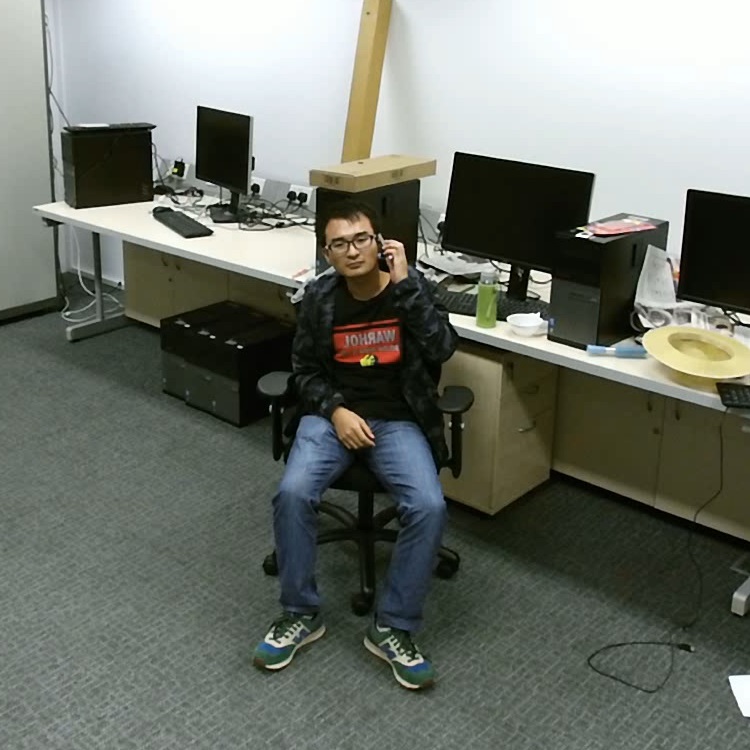} &
		\includegraphics[width=85pt]{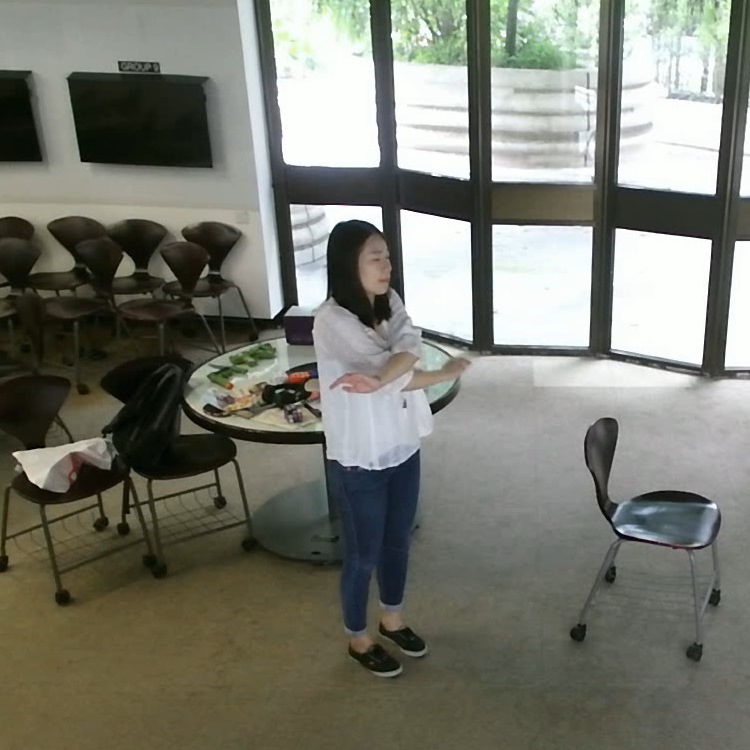} &
		\includegraphics[width=85pt]{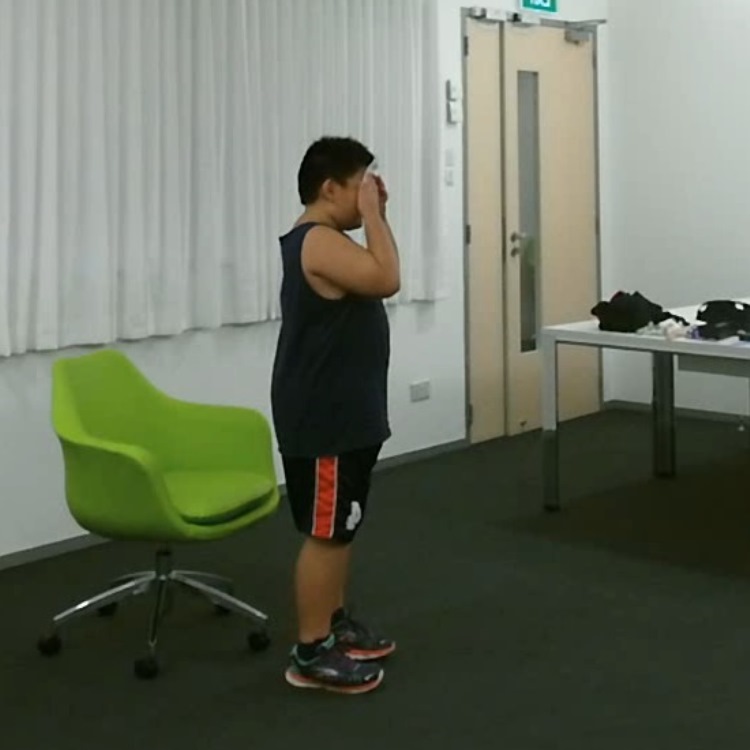} \\
		\includegraphics[width=85pt]{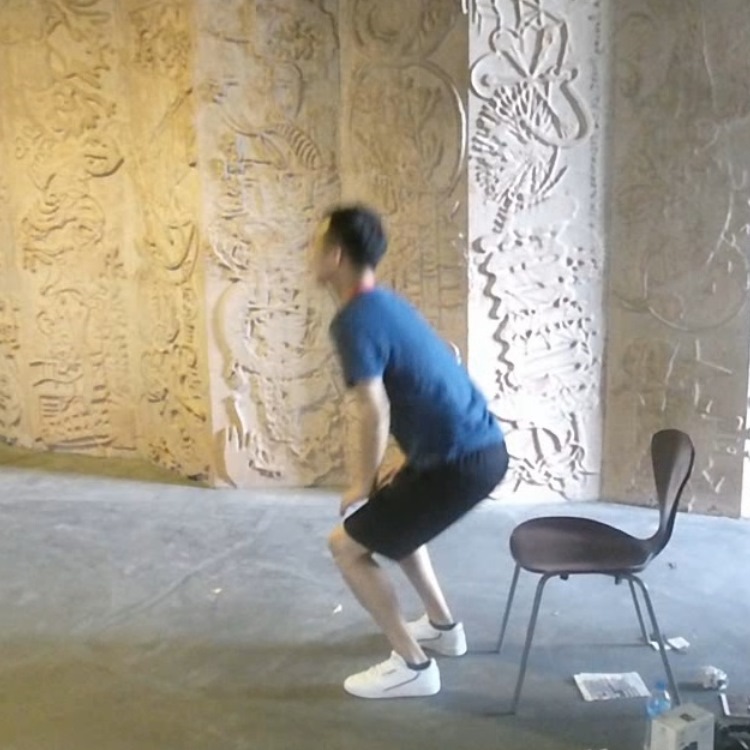} &
		\includegraphics[width=85pt]{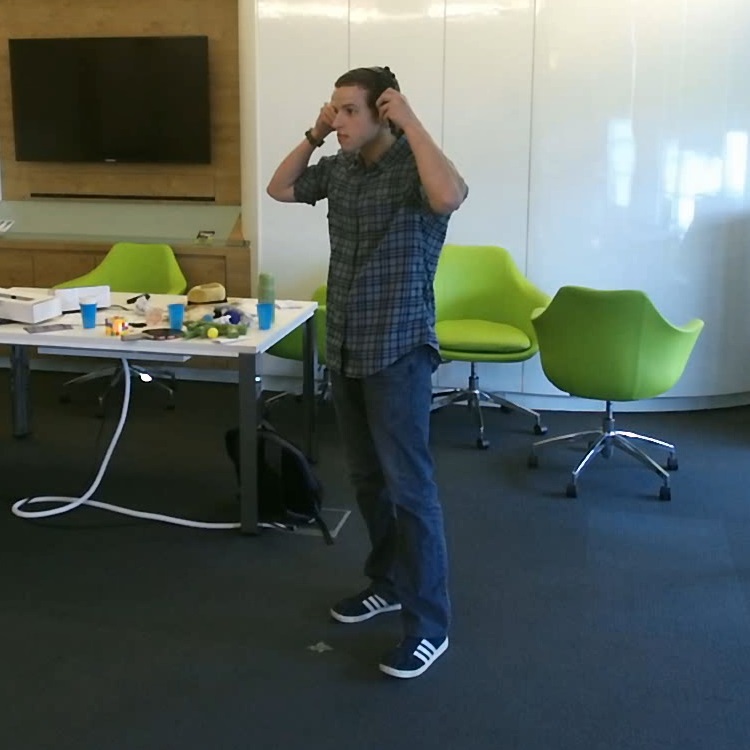} &
		\includegraphics[width=85pt]{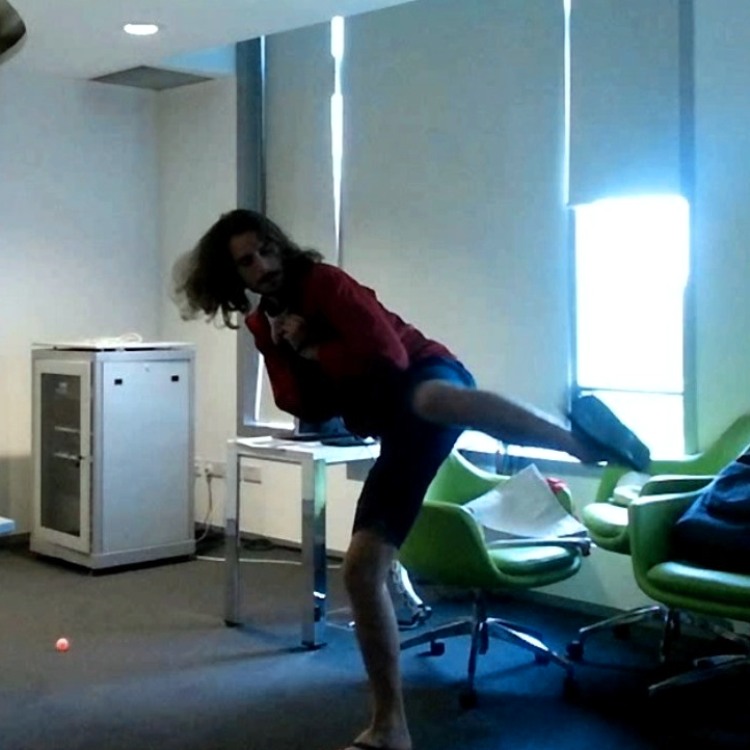} &
		\includegraphics[width=85pt]{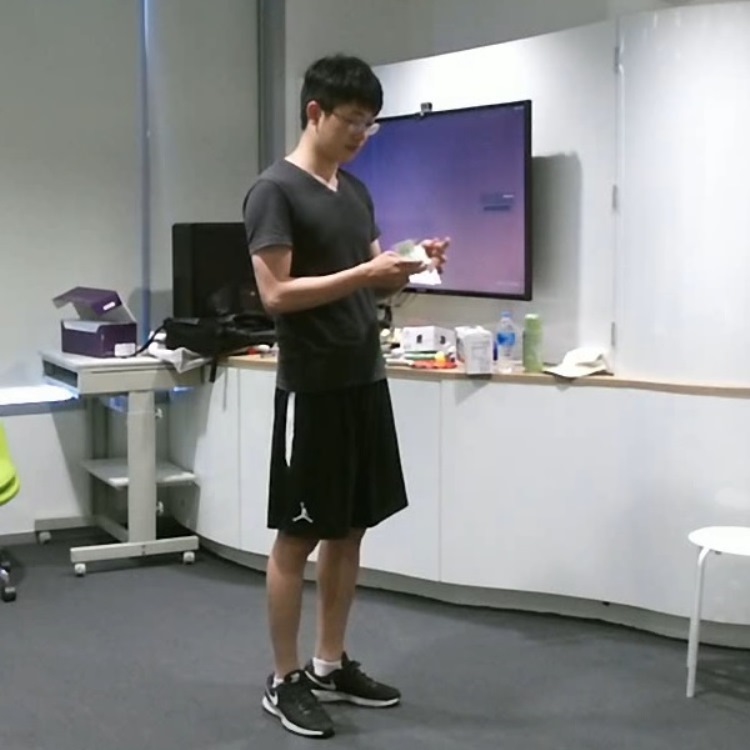} &
		\includegraphics[width=85pt]{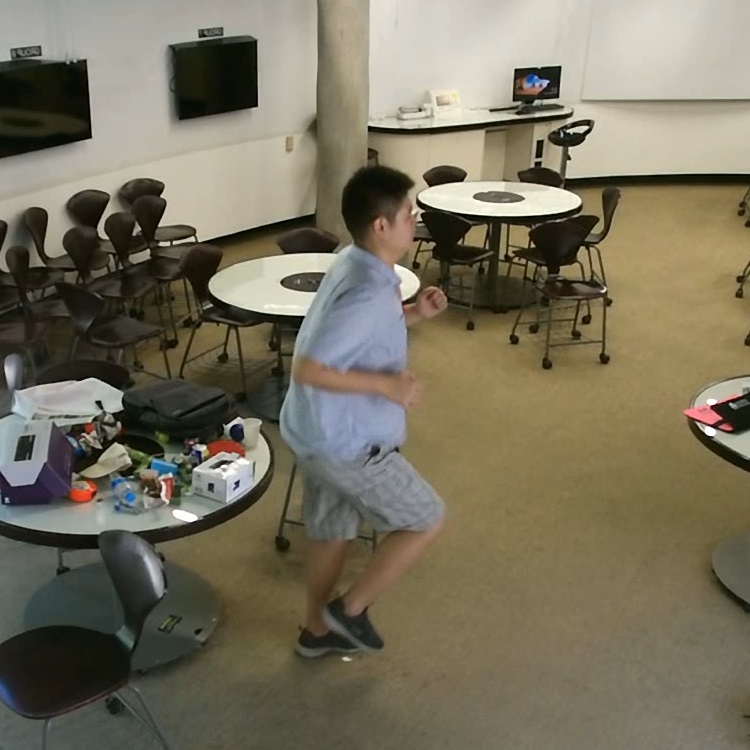} \\
		\includegraphics[width=85pt]{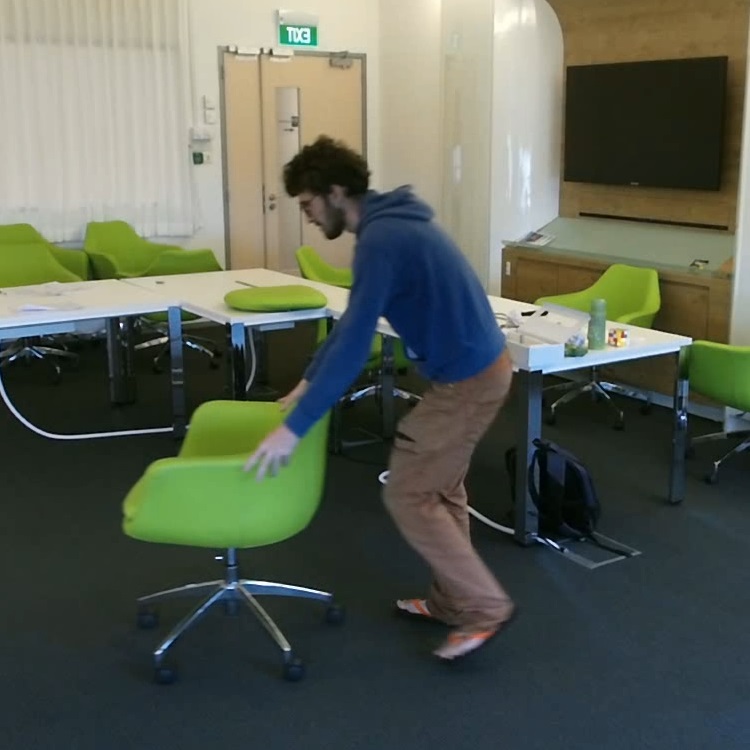} &
		\includegraphics[width=85pt]{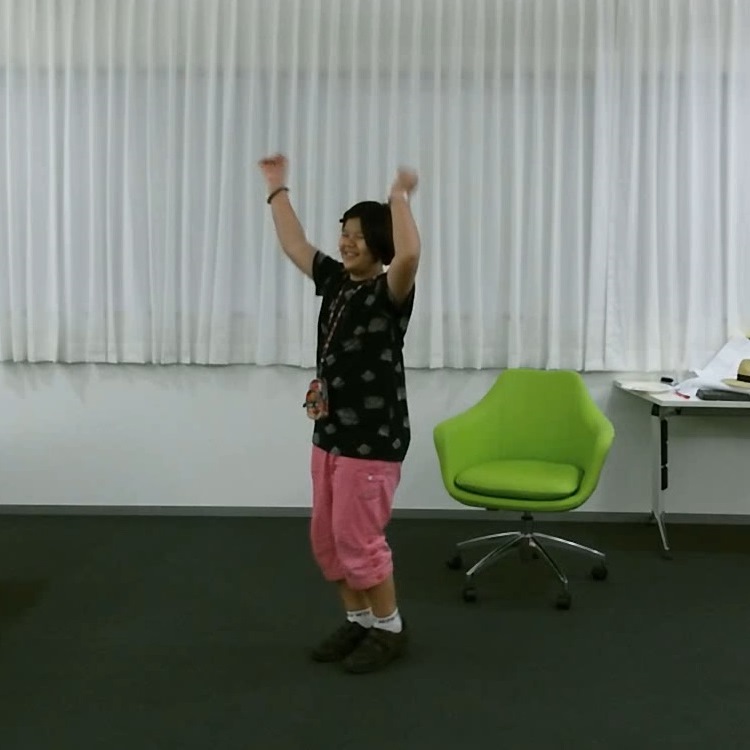} &
		\includegraphics[width=85pt]{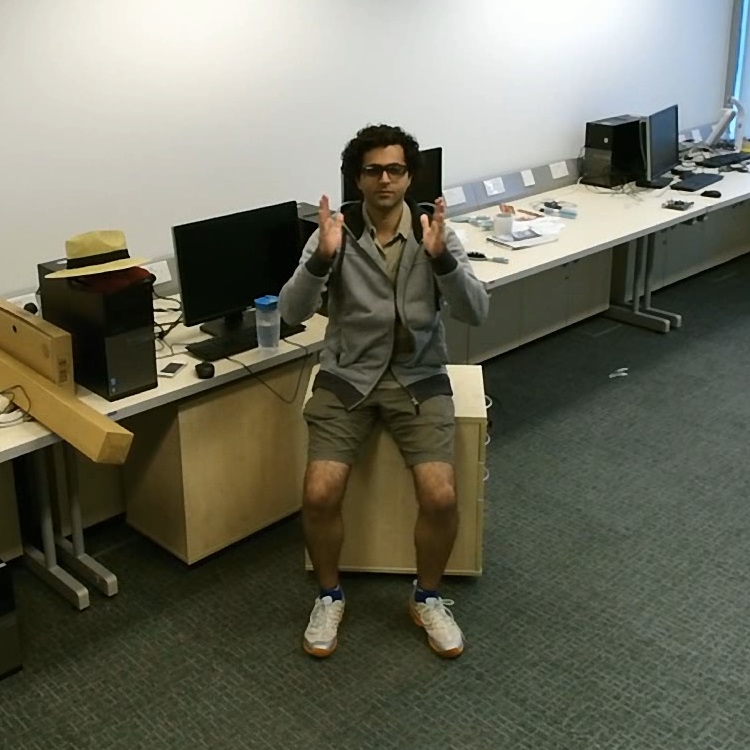} &
		\includegraphics[width=85pt]{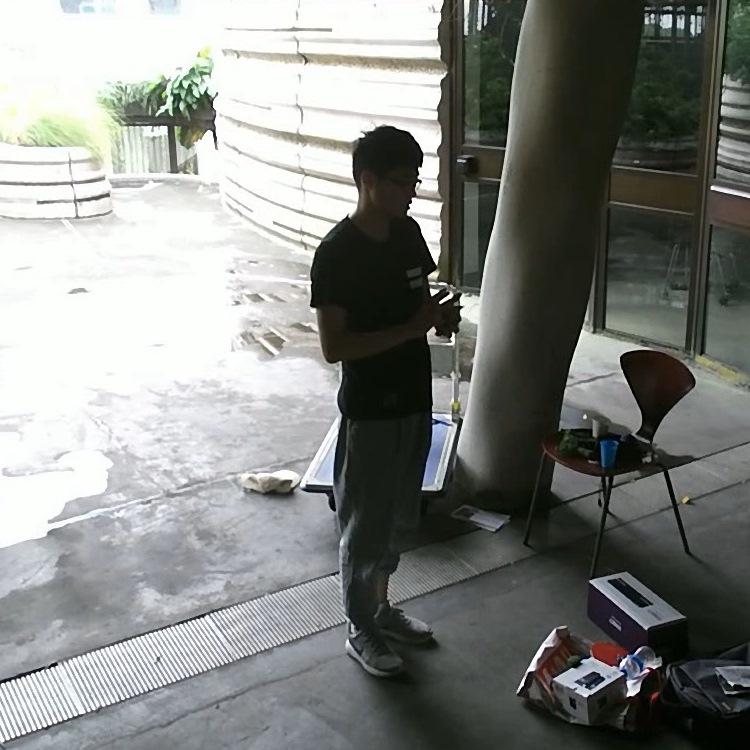} &
		\includegraphics[width=85pt]{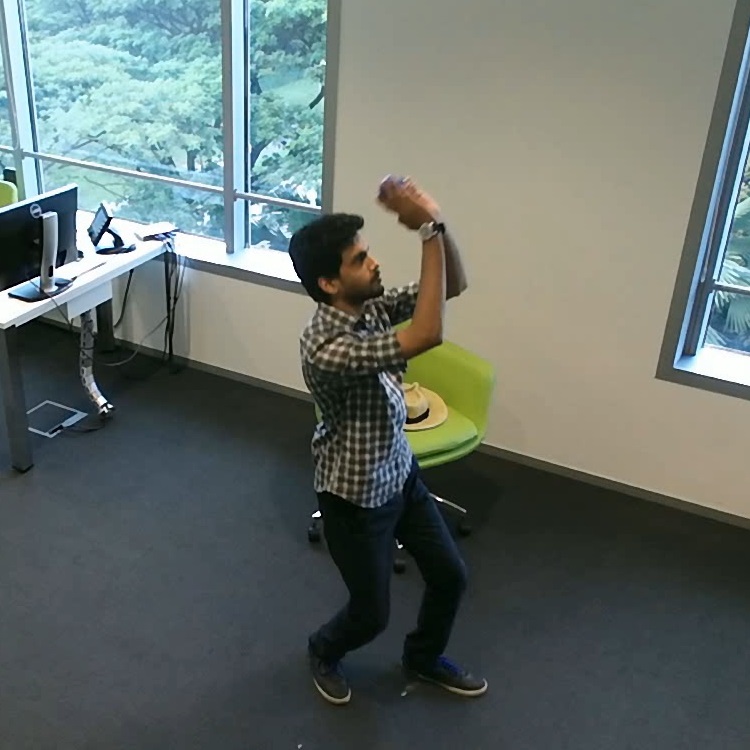} \\
		\includegraphics[width=85pt]{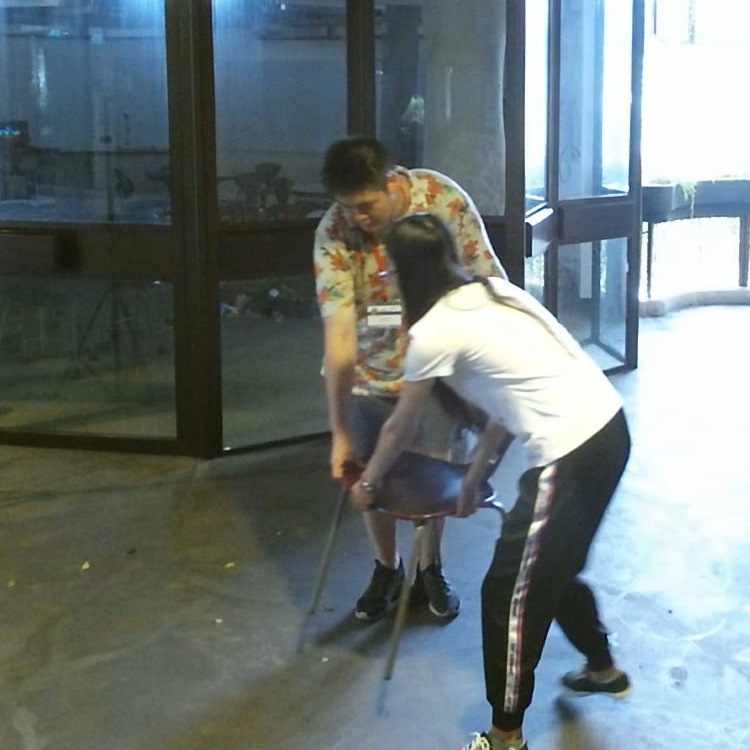} &
		\includegraphics[width=85pt]{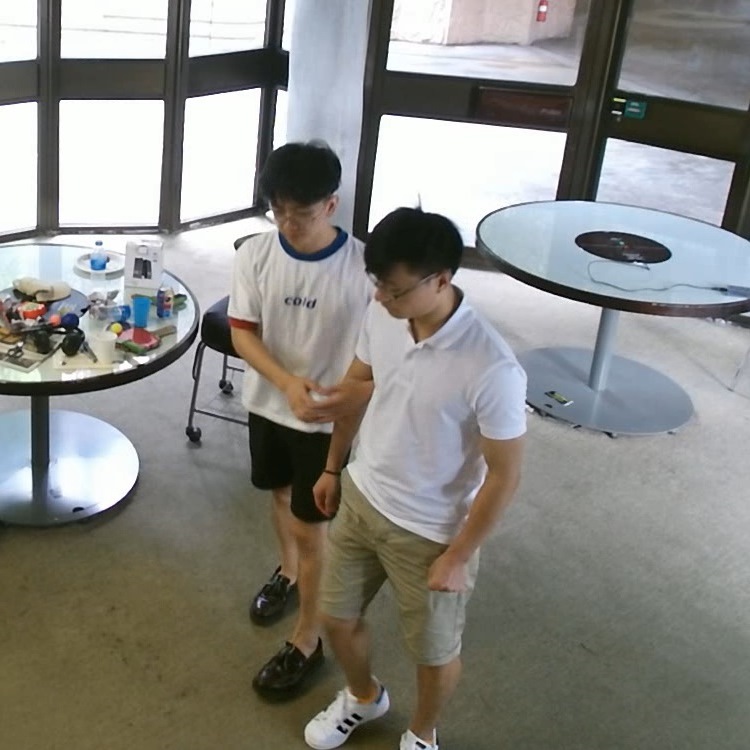} &
		\includegraphics[width=85pt]{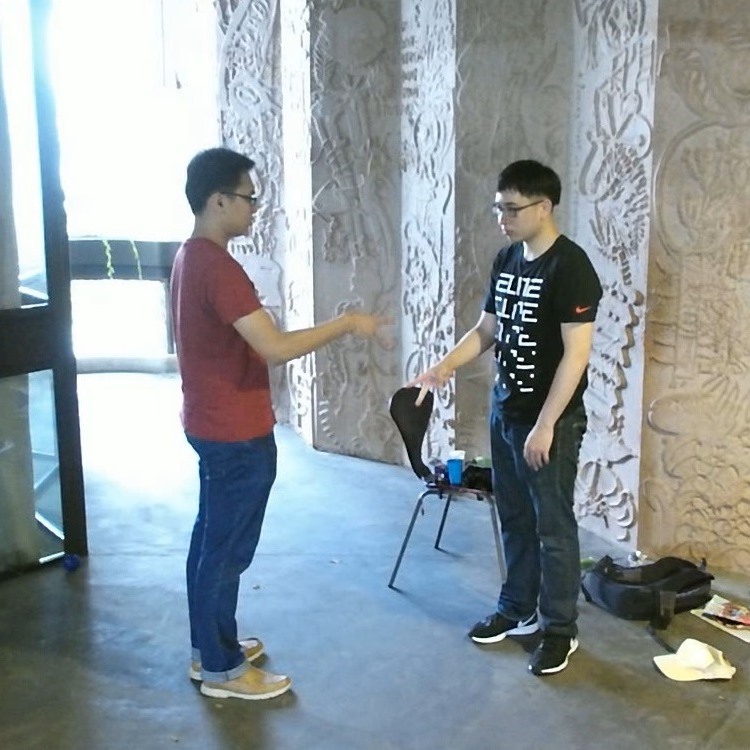} &
		\includegraphics[width=85pt]{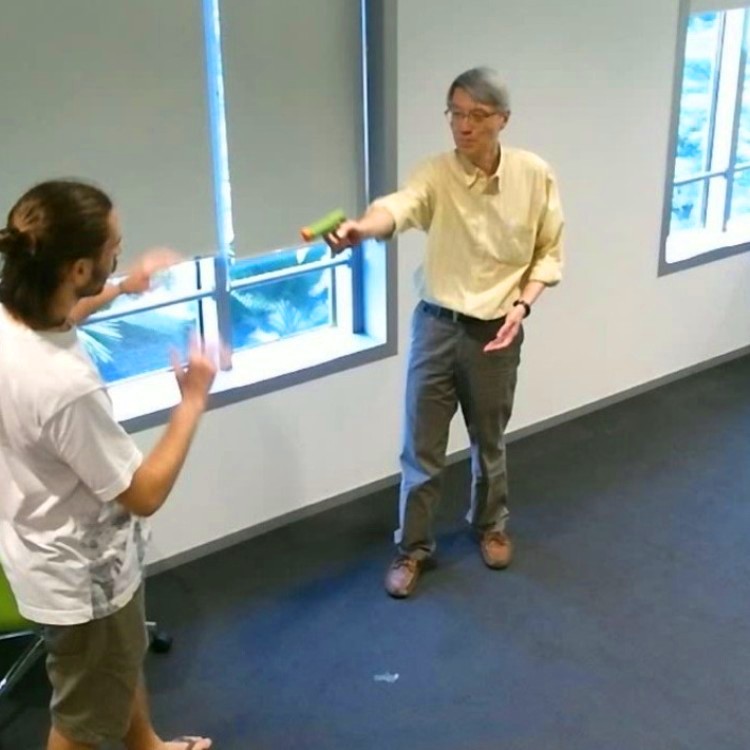} &
		\includegraphics[width=85pt]{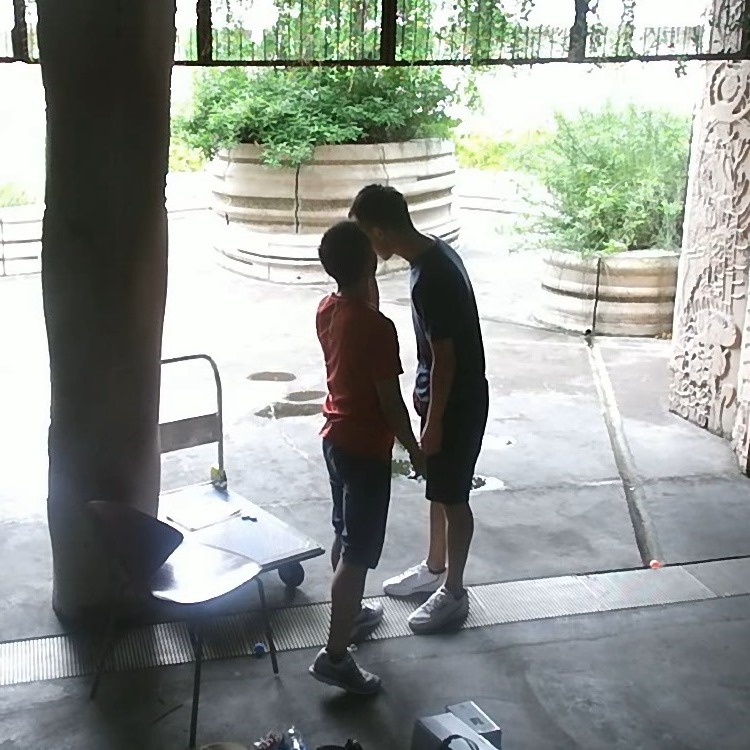} \\&&&&\\
		\includegraphics[width=85pt]{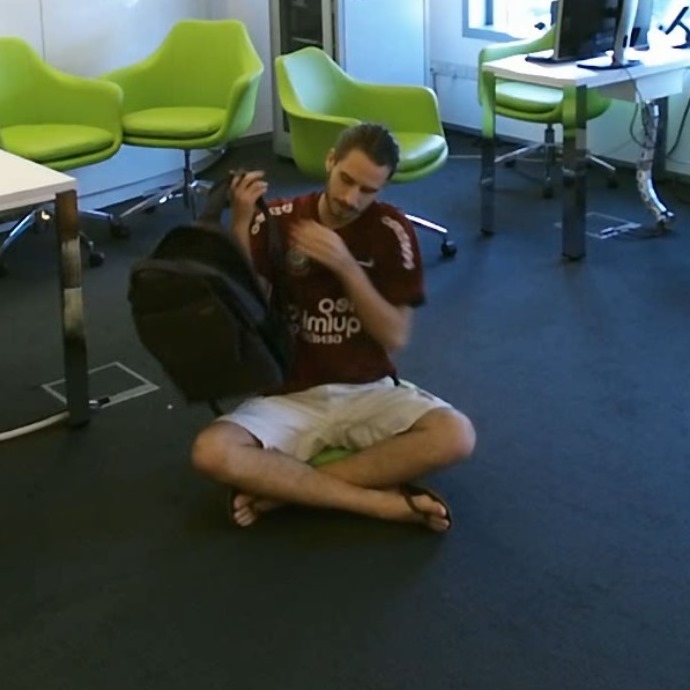} &
		\includegraphics[width=85pt]{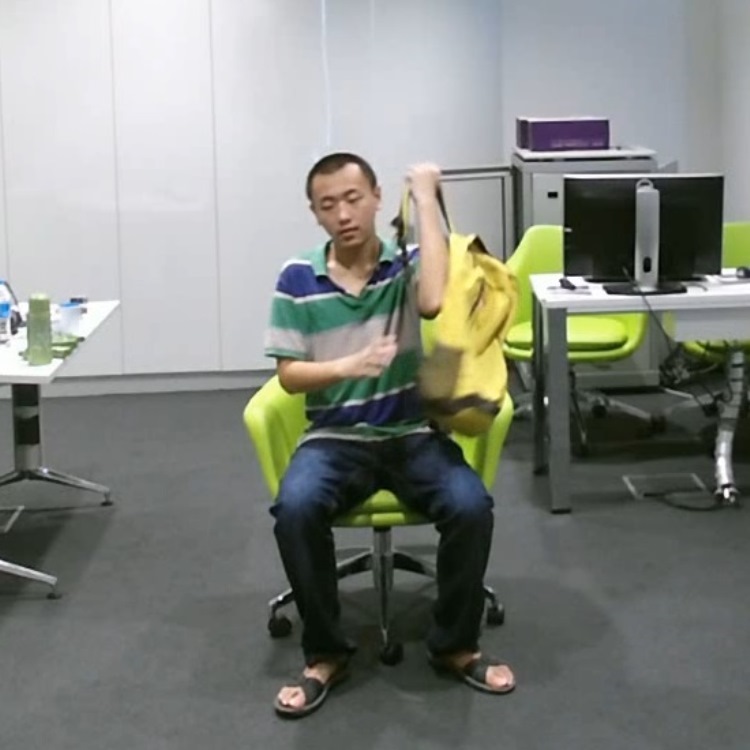} &
		\includegraphics[width=85pt]{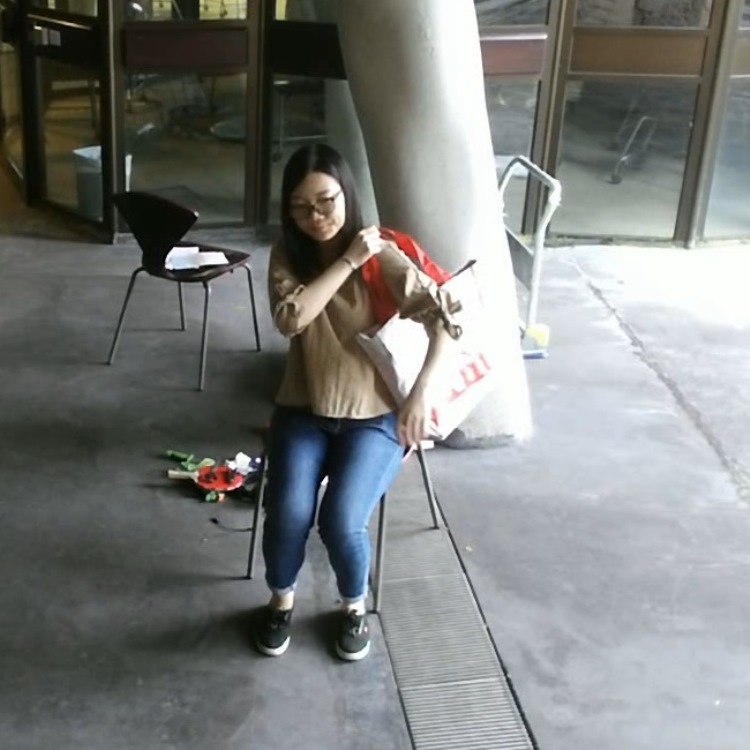} &
		\includegraphics[width=85pt]{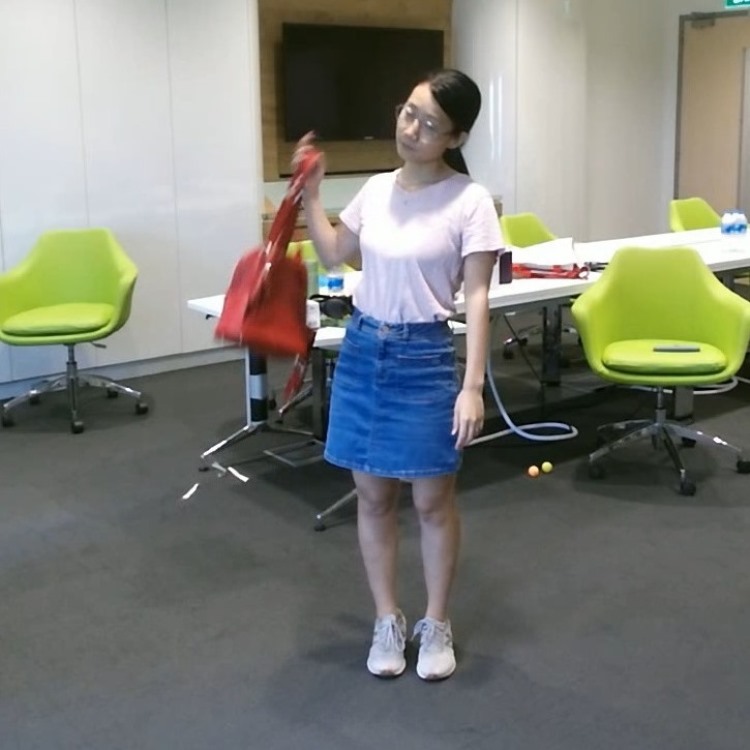} &
		\includegraphics[width=85pt]{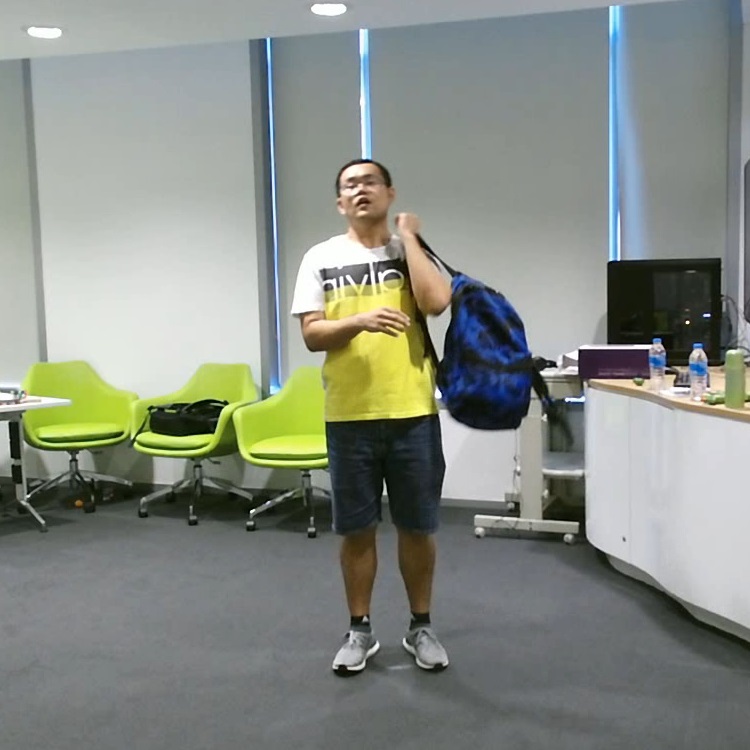} \\&&&&\\
		\includegraphics[width=85pt]{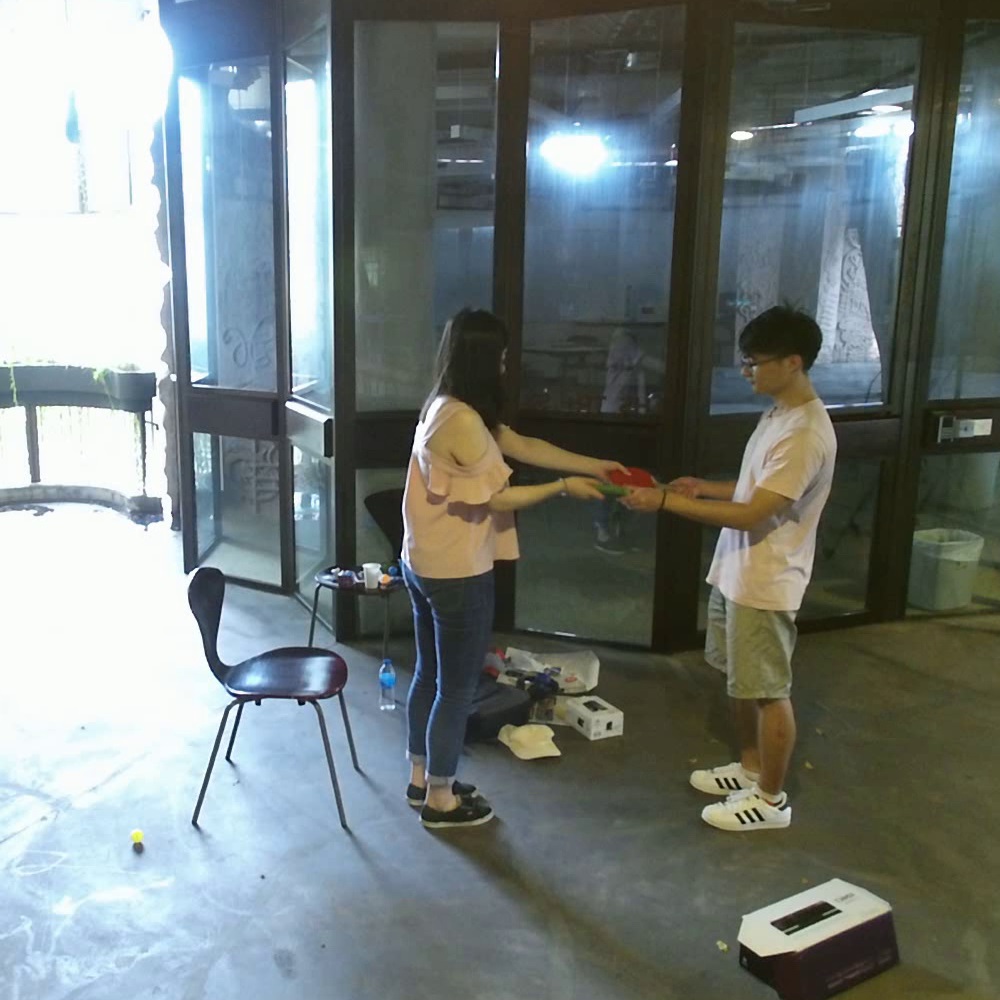} &
		\includegraphics[width=85pt]{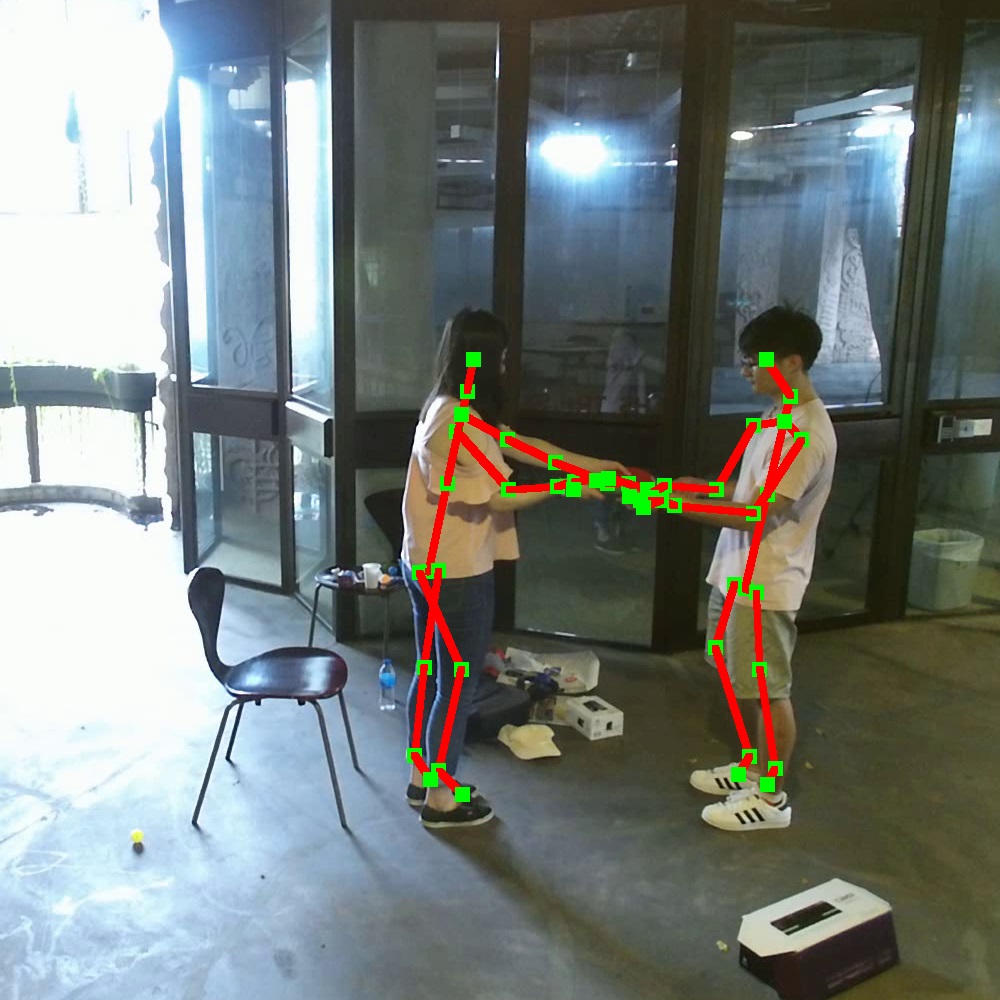} &
		\includegraphics[width=85pt]{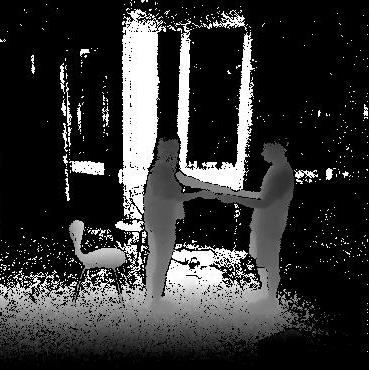} &
		\includegraphics[width=85pt]{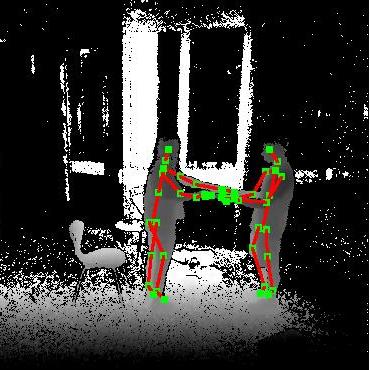} &
		\includegraphics[width=85pt]{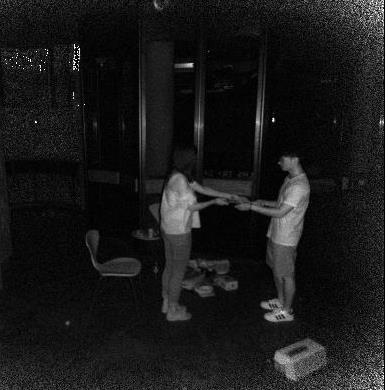} \\
	\end{tabular}
	\vspace{2pt}
	\caption{Sample frames of the NTU RGB+D 120 dataset.
		The first four rows show the variety in human subjects, camera views, and environmental conditions.
		The fifth row depicts the intra-class variation of the performances.
		The last row illustrates the RGB, RGB+joints, depth, depth+joints, and IR modalities of a sample frame.}
	\label{fig:sampleframes}
\end{figure*}

%
%

\ifCLASSOPTIONcaptionsoff
  \newpage
\fi

\bibliographystyle{IEEEtran}
\bibliography{AmirPhDCon}

\end{document}